\documentclass[11pt]{article}

\usepackage[preprint]{acl}

\usepackage{times}
\usepackage{latexsym}

\usepackage{booktabs}   
\usepackage{graphicx}   
\usepackage{pifont}     
\usepackage{multicol}
\usepackage{multirow}
\usepackage{url}

\usepackage{algorithm}
\usepackage{amsmath}

\usepackage{caption}
\usepackage{algpseudocode}
\usepackage[T1]{fontenc}

\usepackage[utf8]{inputenc}

\usepackage{microtype}

\usepackage{inconsolata}

\usepackage{graphicx}

\usepackage{amssymb}

\title{\textsc{AuditFlow}: Executable Symbolic Environments for Structured Financial Reporting Verification}

\author{
 \textbf{Yan Wang\textsuperscript{1}},
 \textbf{Xuguang Ai\textsuperscript{1}},
 \textbf{Jaisal Patel\textsuperscript{2}},
 \textbf{Xueqing Peng\textsuperscript{1}},
\\
 \textbf{Fengran Mo\textsuperscript{3}},
 \textbf{Yupeng Cao\textsuperscript{4}},
 \textbf{Haohang Li\textsuperscript{4}},
 \textbf{Mingyu Cao\textsuperscript{5}},
\\
 \textbf{Lingfei Qian\textsuperscript{1,*}},
 \textbf{Víctor Gutiérrez-Basulto\textsuperscript{6}}
\\
\\
 \textsuperscript{1}The Fin AI, USA,
 \textsuperscript{2}Rensselaer Polytechnic Institute, USA,
 \textsuperscript{3}Université de Montréal, Canada,
\\
 \textsuperscript{4}Stevens Institute of Technology, USA,
 \textsuperscript{5}University of Surrey, UK,
 \textsuperscript{6}Cardiff University, UK
\\
 \small{
   \textbf{Correspondence:} \href{mailto:lfqian94@gmail.com}{lfqian94@gmail.com}
 }
}

\begin{document}
\maketitle
\begin{abstract}
Structured financial reporting verification requires connecting reported facts to accounting constraints, reconstructing expected values, and executing numerical checks. We study this problem through XBRL numerical consistency verification and introduce \textsc{AuditFlow}, a graph-grounded multi-agent framework built around an executable symbolic environment. The environment links a reusable US-GAAP taxonomy graph with filing-specific XBRL evidence and provides typed tools for retrieval, traversal, computation, and rule evaluation. Two specialized auditors investigate constraints and filing evidence, while a senior auditor coordinates disagreement-triggered refinement. Deterministic tools produce verification evidence; evidential aggregation combines structured reports, including agent-declared overrides, into final decisions. On a newly collected 1{,}000-instance evaluation set of engine-recorded DQC violations, \textsc{AuditFlow} achieves $87.70\%$ joint audit accuracy with Claude Sonnet 4.6, compared with $68.60\%$ for Single Agent and $86.70\%$ for Checker-only. Component and coverage analyses identify executable verification as the main source of accuracy, while additional gains from agent judgment depend on the backbone and checker coverage.

\end{abstract}

\section{Introduction}
\label{sec:introduction}

Public companies disclose financial reports that investors, regulators, and auditors use to assess performance and compliance. Many of these reports are filed in eXtensible Business Reporting Language (XBRL), where each reported number is linked to an accounting concept, period, unit, and context. Verifying such a number is not just a matter of finding it in the filing. The system must determine whether it is numerically consistent with the relevant taxonomy constraints and related filing facts.

\begin{figure}
    \centering
    \includegraphics[width=1\linewidth]{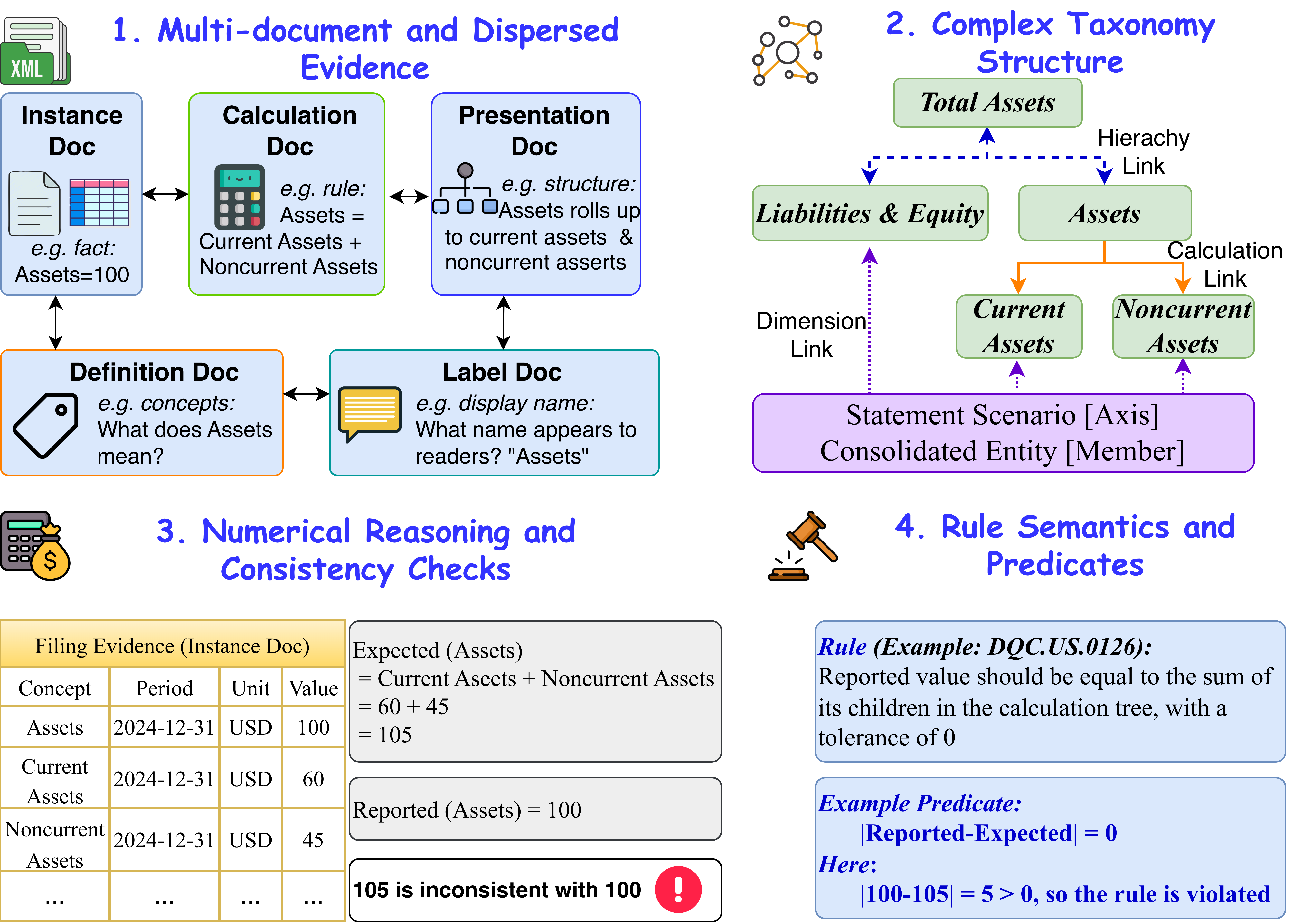}
    \caption{An illustrative calculation-tree verification case. The filing reports \texttt{Assets}=100, while the calculation relation gives an expected value of $60+45=105$ from \texttt{Current Assets} and
    \texttt{Noncurrent Assets}. Because the difference exceeds the rule
    tolerance, DQC.US.0126 is violated.}
    \label{fig:motivating_example}
\end{figure}

This verification setting provides a challenging testbed for financial AI
agents. As shown in Figure~\ref{fig:motivating_example}, deciding whether a
reported value is correct may require linking dispersed XBRL documents,
traversing taxonomy relations, reconstructing an expected value, and applying rules over structured evidence. We model this evidence as a \emph{symbolic environment} comprising explicit entities, relations, constraints, and executable operators. Following the FinMR task introduced by FinAuditing~\citep{10.1145/3805712.3808578}, we study XBRL numerical
consistency verification, where correctness depends on how filing evidence
interacts with taxonomy constraints and calculation structures.

FinAuditing~\citep{10.1145/3805712.3808578} reports only $13.86\%$ accuracy for its strongest evaluated model on numerical consistency verification, illustrating the difficulty of the task in its original evaluation setting. Related challenges arise in legal compliance and clinical guideline checking, where decisions also depend on external rules and structured evidence~\citep{zhang2026guidelinegroundedevidenceaccumulationhighstakes,liu2026evomdt}. However, verification in XBRL also requires exact recomputation of machine-readable facts governed by filing-specific accounting structures. These requirements motivate our central question: \emph{how can language-model agents make reliable decisions through interaction with an executable symbolic environment?}


Prior work grounds LLM reasoning through tools, retrieval, graphs, neuro-symbolic validation, and multi-agent interaction~\citep{yao2023reactsynergizingreasoningacting,schick2023toolformer,patil2024gorilla,edge2024local,amayuelas2025groundingllmreasoningknowledge,feng2025vericotneurosymbolicchainofthoughtvalidation,fatemi2024enhancing,lee2026structureddebateimprovescorporate}. Financial systems such as XBRL-Agent, FinReporting, and Herculean extend these capabilities to filings~\citep{han2024xbrl,zhang2026finreportingagenticworkflowlocalized,peng2026herculeanagenticbenchmarkfinancial}. The challenge is to connect evidence discovery, numerical reconstruction, and rule evaluation over consistent representations of filing facts and reusable accounting constraints. Retrieving relevant evidence alone does not establish numerical correctness.


We propose \textsc{AuditFlow} to separate adaptive evidence search from deterministic computation. Its executable symbolic environment connects a reusable US-GAAP taxonomy graph with filing-specific evidence and performs retrieval, traversal, numerical checks, and rule evaluation. Two junior auditors follow complementary search policies over constraints and filing facts. A senior auditor directs further investigation when their reports disagree; evidential aggregation combines those reports into the final verdict.

We evaluate \textsc{AuditFlow} on 1{,}000 engine-recorded violations covering three DQC rule families, collected using FinAuditing's methodology. With Claude Sonnet 4.6, it reaches $87.70\%$ joint audit accuracy, compared with $68.60\%$ for Single Agent and $86.70\%$ for Checker-only. The latter comparison shows a modest gain over deterministic verification. Removing the checkers reduces accuracy by $60.00$ points on GPT-6.1-sol and $26.67$ on Sonnet in 300-instance ablations. Overrides recover missed violations for some backbones but hurt others. Access to the environment also improves four coding-agent harnesses by $8$--$12$ points on a difficulty-selected subset.

Our contributions are as follows:
(1) We formulate \textbf{graph-grounded numerical consistency verification} over XBRL filings, where decisions require combining filing facts and taxonomy constraints through executable checks with inspectable evidence.
(2) We introduce an \textbf{executable symbolic audit environment} that connects reusable US-GAAP constraints with filing-specific evidence and performs audit-critical retrieval, computation, and rule evaluation through typed deterministic tools.
(3) We evaluate \textbf{agent--environment interaction} through role ablations, tool constraints, evidential aggregation, and overrides, showing that deterministic checking provides most of the accuracy while additional agent benefits depend on the backbone and checker coverage.

\section{Related Work}
\label{sec:related}

Prior work uses tools, retrieval, graphs, neuro-symbolic validation, and
multi-agent interaction to improve LLM grounding and
verification~\citep{yao2023reactsynergizingreasoningacting,
schick2023toolformer, patil2024gorilla, edge2024local,
amayuelas2025groundingllmreasoningknowledge,
feng2025vericotneurosymbolicchainofthoughtvalidation,
fatemi2024enhancing, lee2026structureddebateimprovescorporate}. Financial
systems and benchmarks extend these techniques to filings and
auditing~\citep{han2024xbrl, wang2025finsage, arun2025finreflectkg,
zhang2026finreportingagenticworkflowlocalized,
peng2026herculeanagenticbenchmarkfinancial,
10.1145/3805712.3808578,
malarkkan2026finrulebenchbenchmarkjointreasoning,
yang2026finvault}. Our focus is their integration for executable verification: graph relations serve as inputs to numerical operators, not only retrieved context. Agents choose investigation paths, deterministic tools produce verification evidence, and required-tool constraints govern ordinary termination. Agent reports and evidential aggregation determine the final decision. A more detailed discussion of related work is provided in Appendix~\ref{app:related-work}.

\section{Methodology}
\label{sec:method}

\textsc{AuditFlow} separates adaptive search from deterministic computation. Agents choose what to inspect, while typed tools retrieve filing facts, traverse taxonomy relations, and execute numerical checks over a dual-graph environment.

\begin{figure*}[t]
    \centering
    \includegraphics[width=1\linewidth]{auditflow_framework.png}
    \caption{Overview of \textsc{AuditFlow}. Tools return deterministic evidence; junior reports may add overrides. The senior directs refinement, while evidential aggregation determines the final verdict.}
    \label{fig:overview}
\end{figure*}

Figure~\ref{fig:overview} shows the four components: the dual graph, its typed tool interface, a three-agent audit protocol, and evidential aggregation. Two junior auditors inspect each case from different views; a senior reviews disagreements and directs further investigation.

\subsection{Task Formulation}
\label{sec:task}

We study \textbf{graph-grounded numerical consistency verification}. Given an XBRL filing $\mathcal{F}$, a target concept $c$, a reporting period $t$, and the corresponding US-GAAP taxonomy release $\mathcal{T}_y$, the system decides whether the reported value $v_{\mathrm{rep}}$ is consistent with the value implied by the taxonomy, filing context, and applicable audit rule.
The system outputs
\begin{equation}
  \big(\hat{y},\; v_{\mathrm{rep}},\; v_{\mathrm{exp}},\;
  \mathcal{R},\; \mathcal{P},\; \mathcal{Z},\; \tau\big),
\end{equation}
where $\hat{y}\in\{\mathrm{consistent},\mathrm{violation}\}$ is the verdict, $v_{\mathrm{exp}}$ the expected value derived from applicable verification checks, $\mathcal{R}$ the reported DQC rules, $\mathcal{P}$ the action path, $\mathcal{Z}$ the supporting evidence, and $\tau$ the committed evidence mass from aggregation. Evaluation scores the numerical artifact, not every output field (Appendix~\ref{app:metrics}).

Following FinAuditing's FinMR formulation~\citep{10.1145/3805712.3808578}, we examine sign consistency (DQC.US.0015), dimensional aggregation (DQC.US.0117), and calculation-tree consistency (DQC.US.0126). The general task is binary, but our evaluation contains engine-recorded violations: it tests numerical reconstruction for known inconsistencies, not false-positive behavior on compliant filings.

\subsection{Dual-Graph Audit Environment}
\label{sec:fkg}

The environment keeps taxonomy knowledge and filing evidence separate. The taxonomy defines the reporting constraints, while the filing provides the reported facts. We represent them as a dual-graph environment:
\begin{equation}
  \mathcal{E}=(\mathcal{G}^{y}_{T}, \mathcal{G}_{F}, \mathcal{A}, O),
\end{equation}
where $\mathcal{G}^{y}_{T}$ is the static taxonomy graph for year $y$, $\mathcal{G}_{F}$ is the filing-specific graph, $\mathcal{A}$ is the typed action space, and $O(a,s)$ is the structured observation returned by action $a$ under audit state $s$. The graph component is
\begin{equation}
  \mathcal{G}=(V_T^y\cup V_F,\; E_T^y\cup E_F\cup E_B),
\end{equation}
where $V$ and $E$ denote each graph's nodes and edges; $E_B$ links facts to their governing taxonomy or extension concepts.

\paragraph{Static regulatory graph.}
The taxonomy graph stores concepts and their metadata, including data type, period type, and balance attribute. Presentation, calculation, and dimensional edges supply the accounting constraints used by the audit tools.

\paragraph{Dynamic filing graph.}
The filing graph stores facts with values, units, decimals, periods, and dimensional contexts, together with filing-specific relations. Calculation checks prioritize filing-specific relationships over taxonomy relationships. Bridge edges support joint traversal of facts and constraints (Appendix~\ref{app:graph-construction}).

\subsection{Typed Action and Observation Interface}
\label{sec:tools}

Each tool has a fixed name, structured arguments, and deterministic implementation. The Compliance Auditor receives the static catalog; the Forensic Auditor receives static and dynamic catalogs. These are tool categories, not exclusive graph access: static tools include fact retrieval and checkers over filing evidence. The senior has no tool access.

\textit{Retrieval} and \textit{traversal} tools obtain facts, metadata, and accounting relations; \textit{forensic} tools compare contexts and related concepts. \textit{Deterministic checkers} execute predefined sign, calculation-tree, and dimensional checks. In the authoritative configuration, agents follow checker outcomes. The main, override-enabled configuration also permits an explicitly declared, evidence-supported violation missed by a checker. This is an agent assertion, not an independently verified checker result; the prompt forbids discarding positive checker findings. Section~\ref{sec:coverage} compares the configurations, and Appendix~\ref{app:tools} lists the tools.

\subsection{Three-Agent Audit Protocol}
\label{sec:agents}

The Compliance Auditor investigates constraints, the Forensic Auditor investigates filing evidence, and the Senior Auditor coordinates disagreement-triggered refinement.

\paragraph{Junior auditors.}
At step $s$, a junior selects action $a_s$, receives observation $o_{s+1}=O(a_s,s)$, and appends it to the evidence trace. Its actions are \texttt{tool\_call}, \texttt{ask\_senior}, and \texttt{finalize}. Finalization returns a structured verdict, triggered rules, reported and expected values, rationale, confidence, and interaction history.

The Compliance Auditor $A_1$ is specification-first. It starts from taxonomy metadata, calculation structure, dimensional axes, and applicable rule families, then checks the reported facts against the identified constraints. The Forensic Auditor $A_2$ is evidence-first. It starts from reported facts, period comparisons, dimensional breakdowns, sibling arithmetic, and unit or decimal consistency, then confirms the result with the same deterministic DQC checks. Thus, $A_1$ and $A_2$ use different search policies over the same environment.

\paragraph{Required-tool gate.}
Ordinary \texttt{finalize} and \texttt{ask\_senior} actions require non-error observations from \texttt{check\_sign}, \texttt{check\_calc\_tree}, and \texttt{check\_\allowbreak dim\_\allowbreak consistency}; $A_2$ must also call \texttt{get\_fact\_history}. Premature actions are rejected with the missing tool names. Budget exhaustion instead forces an output from available observations without completing missing checks. Non-error execution need not yield a conclusive rule evaluation (Appendix~\ref{app:implementation}).

\paragraph{Senior auditor.}
Agreement requires matching binary verdicts and reported rule sets, not matching numerical values. Otherwise, the senior directs further investigation of missing or conflicting evidence. Final aggregation selects a verdict and a primary junior report; its numerical fields are not averaged or independently reconciled (Appendix~\ref{app:fusion}).

\subsection{Evidential Aggregation}
\label{sec:fusion}

Each junior report defines a mass function over violation, consistency, and unresolved evidence. Rule-specific reliabilities assign support from the reported rules, including declared overrides when enabled; the free-form rationale and self-reported confidence do not enter this mapping. When no rule is reported, support for consistency depends on whether the reported value has been resolved. The gate checks tool execution during ordinary termination.

We combine the two mass functions using evidential reasoning~\citep{yang2013evidential}. Conflict mass $K$ records opposing support, and the fused ignorance mass $m^\star(\Theta)$ yields the trustworthiness score
\begin{equation}
\tau=1-m^\star(\Theta).
\end{equation}
The final verdict compares support for violation and consistency after redistributing ignorance. The score measures committed mass under this mapping, not a probability of correctness. Aggregation does not independently validate agent-declared overrides. Appendix~\ref{app:fusion} gives the mapping and fusion equations.

\subsection{Bounded Refinement}
\label{sec:refine}

The protocol permits up to three senior-guided refinement rounds after the initial junior investigation. Agreement ends refinement; unresolved disagreement at the limit proceeds to evidential aggregation. Algorithm~\ref{alg:audit} in Appendix~\ref{app:audit-procedure} specifies the loop.

\section{Experimental Setup}
\label{sec:setup}

\subsection{Dataset Construction}
\label{sec:data}

We construct a new 1{,}000-instance evaluation set following FinAuditing's FinMR collection methodology~\citep{10.1145/3805712.3808578}. It spans 548 filings from 319 issuers, including 10-K and 10-Q reports dated 2021--2026. Each instance is a concept--period case associated with a filing, an engine-recorded DQC violation, and reference values; filings and issuers may contribute multiple cases. We parse each filing and its taxonomy release to build the environment.

From this set, we draw 300 instances for ablations and 100 for the external-agent study, approximately preserving rule proportions. The latter selects structurally demanding cases and does not estimate population performance. Appendix~\ref{app:dataset} reports statistics for all three sets.

\subsection{Models}
\label{sec:models}

We evaluate six backbones: Claude Sonnet 4.6~\citep{anthropic2026claudesonnet46systemcard}, GPT-4o~\citep{hurst2024gpt}, Qwen3.5-397B-A17B~\citep{qwen3.5}, GPT-6.1-sol~\citep{openai2026gpt61solsystemcard}, Qwen3.6-27B~\citep{qwen3.6-27b}, and Fino1-14B~\citep{qian2025fino1}. All LLM-based methods use the same set of backbones. API identifiers and serving configurations are given in Appendices~\ref{app:models} and~\ref{app:implementation}.

\subsection{Baselines}
\label{sec:baselines}

We compare eight baselines. \textbf{Direct LLM}~\citep{brown2020language} predicts from filing context; \textbf{Vanilla RAG}~\citep{lewis2020retrieval}, \textbf{GraphRAG}~\citep{edge2024local}, and \textbf{TreeRAG}~\citep{sarthi2024raptor} retrieve textual, graph, and hierarchical evidence. \textbf{FinAuditing}~\citep{10.1145/3805712.3808578} and \textbf{Herculean}~\citep{peng2026herculeanagenticbenchmarkfinancial} provide financial verification workflows. \textbf{Single Agent}~\citep{yao2023reactsynergizingreasoningacting} retains the Forensic Auditor's search policy and full tool catalog, including the executable environment, but removes the other auditors. \textbf{Checker-only} runs deterministic checks without an LLM. The methods share evaluation instances and applicable backbones, but differ in architecture, evidence access, and execution.

\subsection{Evaluation Metrics}
\label{sec:metrics}

\textbf{Joint audit accuracy} (Joint ACC) scores usable outputs with correct reported and expected values, following FinAuditing~\citep{10.1145/3805712.3808578}. Structural, extraction, and calculation errors (SER, EER, CER) partition failures. \textbf{Verdict-only accuracy} (VAcc) diagnoses recognition of known violations; it does not measure specificity on compliant filings. The fixed \texttt{gpt-5-mini} judge uses cached labels. Every main run has a 1{,}000-instance denominator, including context-window failures; judge failures are retried.

Two co-authors with expertise in finance independently reviewed 100 Claude Sonnet 4.6 \textsc{AuditFlow} outputs, agreeing on $97/100$ cases ($97.0\%$; Cohen's $\kappa\approx0.89$). After disagreement resolution, their consensus matched the judge on $98/100$ cases ($98.0\%$; $\kappa\approx0.94$). Both experts followed the same four-category annotation guidelines, with the complete instructions and examples provided in Appendix~\ref{app:human-annotation}.

Rule attribution affects numerical value selection, but is not independently scored. Main runs use system-identified rules without reference labels; Appendices~\ref{app:metrics} and~\ref{app:protocol} detail scoring and the oracle control.

\subsection{Implementation Details}
\label{sec:impl-details}

All three auditors use the same backbone within a run. We use temperature $0$ where supported and omit it for reasoning models that do not accept the parameter. Each junior invocation has at most $20$ ReAct steps, with up to $3$ refinement rounds after the initial investigation. The senior reviews junior reports without tool access. Main experiments use the expanded sign checker and allow agents to flag evidence-supported violations missed by a silent checker; Section~\ref{sec:coverage} varies both settings. Model-specific output budgets, local serving details, and caching are given in Appendix~\ref{app:implementation}.

\section{Results}
\label{sec:results}

\subsection{Main Comparison with Baselines}
\label{sec:main-comparison}

\begin{figure*}[t]
\centering
\includegraphics[width=\linewidth]{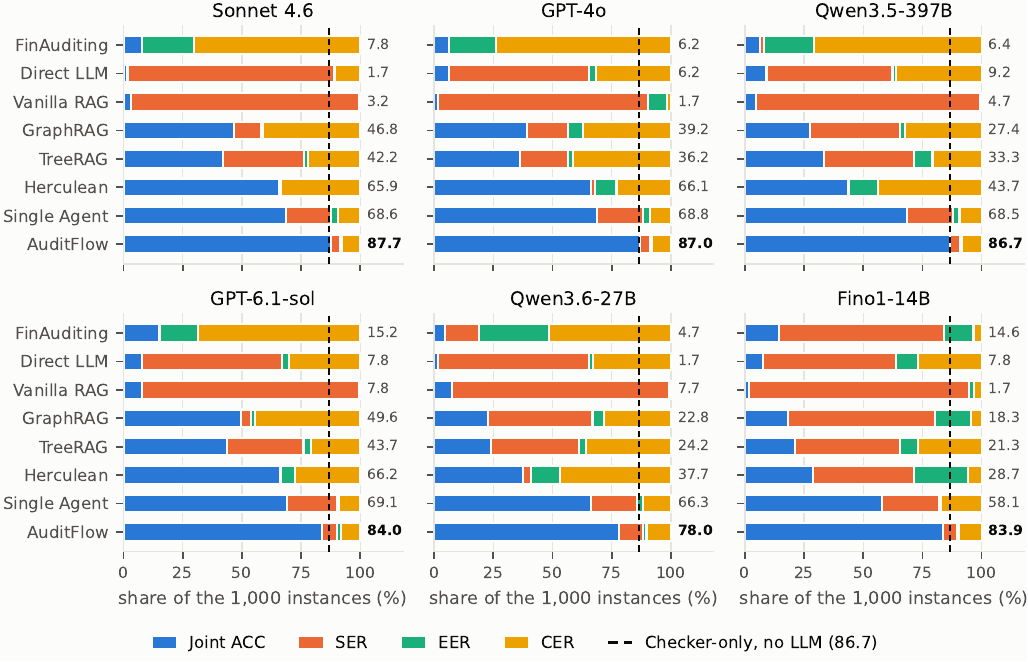}
\caption{Main results across six backbones on the same 1{,}000 instances. Each bar partitions the sample into Joint ACC, structural errors (SER), extraction errors (EER), and calculation errors (CER). Numbers beside the bars report Joint ACC (\%). The dashed line marks Checker-only accuracy ($86.70\%$), obtained without an LLM. All systems select their reported values using their own rule attribution. Tables~\ref{tab:full} and~\ref{tab:full-cont} provide the complete numerical results.}
\label{fig:main-results}
\end{figure*}

\begin{figure*}[t]
\centering
\includegraphics[width=\linewidth]{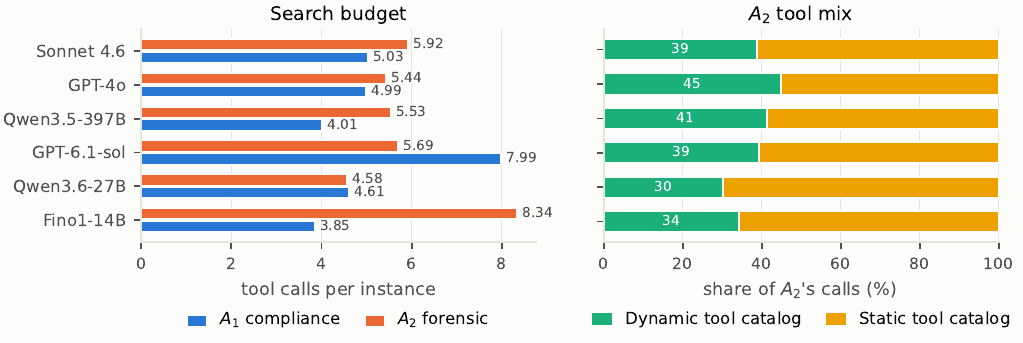}
\caption{Tool use by the two junior auditors. Left: mean tool calls per instance for the Compliance Auditor ($A_1$) and Forensic Auditor ($A_2$). Right: the shares of $A_2$ calls using the dynamic and static tool catalogs. The labeled dynamic shares are rounded to whole percentages. Statistics use completed trajectories: 1{,}000 per backbone except Fino1-14B, with 997. Exact values are in Table~\ref{tab:tooluse}.}
\label{fig:tooluse}
\end{figure*}

\begin{figure}[t]
\centering
\includegraphics[width=\linewidth]{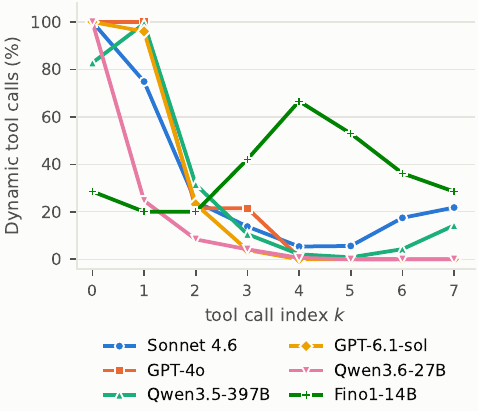}
\caption{Temporal tool-use profile of the Forensic Auditor $A_2$. Each point gives the share of dynamic filing-tool calls at zero-indexed step $k$. Five backbones begin primarily with filing evidence and shift toward taxonomy tools and deterministic checks; Fino1-14B gathers more filing evidence later. Later steps include only trajectories that reach them.}
\label{fig:switchover}
\end{figure}

Figure~\ref{fig:main-results} compares \textsc{AuditFlow} with the baselines across six backbones. \textsc{AuditFlow} achieves the highest Joint ACC among LLM-based methods under every backbone, ranging from $78.00\%$ to $87.70\%$. With Sonnet, it scores $87.70\%$, versus $68.60\%$ for Single Agent and $65.90\%$ for Herculean. The Single Agent comparison tests the additional audit protocol with tools already available; the other baselines also differ in evidence access and execution.

Checker-only provides an important qualification: it already reaches $86.70\%$. Relative to this baseline, Sonnet improves by $1.00$ point, with 11 corrected cases and one regression (two-sided exact McNemar, $p=0.006$). GPT-4o improves by $0.30$ points (4 wins, 1 loss; $p=0.375$), while Qwen3.5-397B-A17B ties the baseline. The other three backbones perform below it. Thus, most of the accuracy comes from the executable environment. The main comparison supports a modest additional gain for Sonnet, rather than a general accuracy benefit from adding a multi-agent loop to well-covered rules.

The error decomposition in Figure~\ref{fig:main-results} also distinguishes different failure modes (full values in Appendix~\ref{app:full-results}). Under Sonnet, Vanilla RAG has $96.10\%$ SER, while Herculean has $33.60\%$ CER. Single Agent still has $19.10\%$ SER, so agentic methods do not uniformly fail only on numerical reconstruction. \textsc{AuditFlow} reduces these rates to $3.60\%$ SER, $0.80\%$ EER, and $7.90\%$ CER; expected-value reconstruction accounts for the largest share of its remaining errors.

\subsection{Backbone Robustness}
\label{sec:backbone-results}

VAcc ranges from $90.60\%$ to $94.80\%$, exceeding Joint ACC by $6.80$--$12.60$ points (Appendix~\ref{app:backbone}). A correct binary decision therefore does not ensure a correct audit artifact. For example, GPT-6.1-sol has the highest VAcc ($94.80\%$), but its Joint ACC ($84.00\%$) is below Sonnet's. The gap reflects numerical extraction and reconstruction errors, including rule-dependent value selection, which VAcc does not test.

Fino1-14B reaches $83.90\%$ Joint ACC and $91.00\%$ VAcc when served without a reasoning parser and with its 32{,}768-token context window. Three context-overflow cases remain in the denominator as failures. Under these serving configurations, it exceeds Qwen3.6-27B's $78.00\%$; parameter count alone does not explain performance. Dimensional aggregation is harder within this evaluation: Joint ACC is $59.15$--$66.20\%$, compared with $76.07$--$78.92\%$ for calculation-tree consistency. Bootstrap intervals and paired comparisons are provided in Appendix~\ref{app:bootstrap}.

\subsection{Ablation Study}
\label{sec:ablation}

\begin{table}[t]
\centering
\footnotesize
\setlength{\tabcolsep}{2.5pt}
\begin{tabular}{lrrrr}
\toprule
\multirow{2}{*}{Variant} & \multicolumn{2}{c}{GPT-6.1-sol} & \multicolumn{2}{c}{Sonnet 4.6} \\
\cmidrule(lr){2-3}\cmidrule(lr){4-5}
 & ACC & $\Delta$ & ACC & $\Delta$ \\
\midrule
Full system  & 83.33 & --- & 88.67 & --- \\
No tool gate    & 80.00 & $-3.33$ & 88.67 & $0.00$ \\
No ER fusion             & 80.67 & $-2.67$ & 88.67 & $0.00$ \\
$A_1$ only    & 78.67 & $-4.67$ & 88.33 & $-0.33$ \\
$A_2$ only    & 70.67 & $-12.67$ & 69.67 & $-19.00$ \\
No checks  & 23.33 & $\mathbf{-60.00}$ & 62.00 & $-26.67$ \\
\bottomrule
\end{tabular}
\caption{Component ablations on the 300-instance sample. ACC is a percentage; $\Delta$ is the change in percentage points. The $A_1$-only and $A_2$-only variants retain the indicated junior; No checks removes the deterministic checkers. Changes are computed before rounding.}
\label{tab:ablation}
\end{table}

Table~\ref{tab:ablation} reports component ablations under GPT-6.1-sol and Sonnet 4.6. Removing deterministic checks causes the largest drop: Joint ACC falls from $83.33\%$ to $23.33\%$ on GPT-6.1-sol and from $88.67\%$ to $62.00\%$ on Sonnet, corresponding to decreases of $60.00$ and $26.67$ points, respectively. These results support the role of deterministic tools in executing verification over structured evidence, with a larger impact on GPT-6.1-sol.

The two junior roles contribute asymmetrically. Retaining only the Compliance Auditor $A_1$ scores $78.67\%$ versus $83.33\%$ for the full system on GPT-6.1-sol and nearly matches it on Sonnet ($88.33\%$ versus $88.67\%$), whereas retaining only the Forensic Auditor $A_2$ reduces accuracy to $70.67\%$ and $69.67\%$, respectively. These variants retain different catalogs and search policies, so the gap does not isolate role specialization. Nor does it establish that two auditors are generally necessary.

Removing the required-tool gate or ER fusion reduces Joint ACC on GPT-6.1-sol by $3.33$ and $2.67$ points, respectively, while leaving accuracy unchanged on Sonnet. These results suggest that their contributions depend on the backbone rather than providing uniform accuracy gains. The components nevertheless serve distinct roles in enforcing the tool-use protocol and consolidating evidence.

\subsection{Checker Coverage and Agent Judgment}
\label{sec:coverage}

DQC.US.0015 prohibits negatives for specified concepts. Our limited checker tests a curated subset, then checks only debit-balance concepts outside it. Credit-balance concepts are exempt because negatives can be valid, as with accumulated deficit. Expanded coverage includes them at a potential precision cost; neither variant reproduces the full official engine. We vary checker coverage and permission for agent-declared overrides separately on the same 300 instances (Appendix~\ref{app:coverage}).

\begin{table}[!htbp]
\centering
\small
\setlength{\tabcolsep}{3pt}
\resizebox{\linewidth}{!}{%
\begin{tabular}{llrrr}
\toprule
Backbone & Coverage & \multicolumn{2}{c}{Joint ACC} & $\Delta$ \\
\cmidrule(lr){3-4}
 & & authoritative & override & \\
\midrule
\multirow{2}{*}{Claude Sonnet 4.6} & limited & 69.67 & \textbf{84.33} & $+14.67$ \\
                             & expanded & 86.67 & \textbf{88.67} & $+2.00$ \\
\midrule
\multirow{2}{*}{GPT-4o}      & limited & 62.67 & \textbf{73.67} & $+11.00$ \\
                             & expanded & 86.67 & 86.67 & $0.00$ \\
\midrule
\multirow{2}{*}{Qwen3.6-27B} & limited & 68.33 & 59.67 & $-8.67$ \\
                             & expanded & 80.00 & 77.67 & $-2.33$ \\
\bottomrule
\end{tabular}}
\caption{Joint ACC (\%) on the same 300 instances. $\Delta$ is the paired gain from allowing overrides, in percentage points. Conditions are detailed in Appendix~\ref{app:coverage}.}
\label{tab:coverage}
\end{table}

Table~\ref{tab:coverage} shows that overrides improve Sonnet and GPT-4o by $14.67$ and $11.00$ points under limited coverage, but only $2.00$ and $0.00$ under expanded coverage. Qwen3.6-27B instead loses $8.67$ and $2.33$ points, respectively. Agent judgment can recover missed violations, but its benefit depends on the backbone and checker coverage.

\subsection{Transfer to General Coding Agents}
\label{sec:external}

We compare four coding-agent harnesses with (\textsc{env}) and without (\textsc{bare}) the environment on 100 difficulty-selected instances. Each pair uses the same harness, model, filing, and rule descriptions; scoring requires exact matches for reported and expected values. Appendix~\ref{app:external} details the protocol.

\begin{table}[!htbp]
\centering
\footnotesize
\setlength{\tabcolsep}{4pt}
\begin{tabular}{@{}rrrrrrr@{}}
\toprule
\textsc{bare} & \textsc{env} & $\Delta$ & Win & Lose & Tie & $p$ \\
\midrule
\multicolumn{7}{@{}l}{\href{https://code.claude.com/docs/en/overview}{\textit{Claude Code}} (Claude Opus 5.5)} \\
83 & \textbf{91} & $+8$ & 10 & 2 & 88 & 0.039 \\
\midrule
\multicolumn{7}{@{}l}{\href{https://developers.openai.com/codex/cli/}{\textit{Codex}} (GPT-6-astra)} \\
85 & \textbf{93} & $+8$ & 9 & 1 & 90 & 0.021 \\
\midrule
\multicolumn{7}{@{}l}{\href{https://github.com/openclaw/openclaw}{\textit{OpenClaw}} (Gemini 3.8 Flash)} \\
79 & \textbf{90} & $+11$ & 13 & 2 & 85 & 0.007 \\
\midrule
\multicolumn{7}{@{}l}{\href{https://deepseek.com/en/harness/}{\textit{DeepSeek harness}} (DeepSeek Flash)} \\
76 & \textbf{88} & $+12$ & 15 & 3 & 82 & 0.008 \\
\bottomrule
\end{tabular}
\caption{Coding agents with and without the environment on 100 difficulty-selected instances. Scores are exact-match accuracy (\%); $\Delta$ is the gain in percentage points. Win/Lose/Tie count paired outcomes; $p$-values are unadjusted exact McNemar tests. Appendix~\ref{app:external} details the protocol.}
\label{tab:external}
\end{table}

Table~\ref{tab:external} shows gains of $8$--$12$ points across all four harnesses (unadjusted exact McNemar $p<0.05$). Neither condition solves the seven dimensional cases. The gains support utility beyond our audit protocol; absolute scores are not comparable with the main evaluation because sampling and scoring differ.

\subsection{Agent Behavior}
\label{sec:agent-behavior}

Figures~\ref{fig:tooluse} and~\ref{fig:switchover} summarize the assigned search policies. The Compliance Auditor uses almost exclusively the static catalog, while the Forensic Auditor mixes static and dynamic tools. Five backbones begin primarily with filing evidence and shift toward taxonomy tools and checks; Fino1-14B uses more filing tools later. These are descriptive patterns under assigned catalogs and checks, not evidence of unconstrained or optimal search.

\paragraph{Role specialization control.}
With GPT-6.1-sol, we give $A_1$ all nine dynamic tools while fixing its persona, required checks, and protocol on the same 300-instance ablation subset. Table~\ref{tab:role-specialization-control} shows that it still makes $99.8\%$ static calls: only four of 2{,}384 calls are dynamic. Calls per instance remain similar ($7.95$ versus $7.97$), as does system Joint ACC ($84.00\%$ versus $83.33\%$). The accuracy difference is only two cases; tool descriptions and mandatory checks remain possible influences on behavior. Thus, static-only access is not necessary for this predominantly specification-first behavior. Together with $A_2$'s access to both catalogs, the result suggests that division of labor is not solely imposed by catalog restrictions; it does not isolate persona effects (Appendix~\ref{app:tool-use}).

\begin{table}[!htbp]
\centering
\footnotesize
\setlength{\tabcolsep}{3pt}
\begin{tabular}{@{}lrrrr@{}}
\toprule
$A_1$ Catalog & Calls & Static & Dynamic & Joint ACC \\
\midrule
Static only & 7.97 & 100.0 & 0.0 & 83.33 \\
Static + dynamic & 7.95 & 99.8 & 0.2 & 84.00 \\
\bottomrule
\end{tabular}
\caption{Catalog-expansion control with GPT-6.1-sol on the same 300 instances. Only $A_1$'s catalog changes; its persona, required checks, and protocol remain fixed. Calls are per instance; shares and system-level Joint ACC are percentages.}
\label{tab:role-specialization-control}
\end{table}

Disagreement is also backbone-dependent: Sonnet's juniors disagree on only 8 cases, while Qwen3.6-27B disagrees on 418, with a smaller accuracy gap between agreement and disagreement cases. Thus, disagreement is not a reliable difficulty signal across backbones. Appendix~\ref{app:agent-behavior} provides the detailed statistics.

\section{Conclusion}
\label{sec:conclusion}

We introduced \textsc{AuditFlow}, whose executable symbolic environment connects XBRL evidence with accounting constraints through typed verification tools. On 1{,}000 engine-recorded violations, it reaches $87.70\%$ Joint ACC with Sonnet, compared with $86.70\%$ for Checker-only. Deterministic verification provides most of the accuracy; additional agent benefits depend on the backbone and checker coverage. The environment also improves four coding-agent harnesses on a difficulty-selected sample. These findings support separating adaptive evidence exploration from executable numerical verification.

\section*{Limitations}
\label{sec:limitations}

Our study focuses on numerical consistency verification for three DQC rule families using engine-recorded XBRL violations. This setting enables controlled evaluation of structured evidence retrieval and numerical reconstruction, but does not assess performance on compliant filings or the broader range of XBRL validation tasks. AuditFlow relies on deterministic checkers implementing selected rule conditions, and extending the framework to additional rules and filing contexts may require further tool development and validation.

Our evaluation covers multiple model backbones and agent configurations, with expert assessment of a subset of outputs. The reported results reflect the evaluated instances and execution settings; broader deployment would benefit from additional validation across filings, rule families, and operational environments. The external coding-agent study uses a difficulty-selected subset and a distinct evaluation protocol, so its scores are not directly comparable with the main evaluation.

\section*{Ethical Considerations}
\label{sec:ethics}

\textsc{AuditFlow} is designed as a research system for XBRL audit verification, not as a replacement for professional audit judgment. Because the task concerns financial filings, incorrect outputs could create risk if users treat the system as an authoritative auditor. A false negative may miss an inconsistency, while a false positive may raise unnecessary concern about a valid filing. We reduce this risk by requiring evidence trails, deterministic rule checks, and explicit reported and expected values in the output, so that each verdict can be inspected rather than accepted as a model-only judgment.

The system uses public-company filing data and engine-recorded rule violations. It does not trade, execute financial actions, or make regulatory decisions. Any real-world use would require additional validation, human review, and integration with existing audit and compliance procedures.

\section*{Artifact Use and Licensing}
\label{sec:artifact-license}

Our experiments use public-company filings retrieved from SEC EDGAR, instance metadata collected following FinAuditing's methodology, and publicly available US-GAAP taxonomy releases. We use these resources for research evaluation and cite their original sources. We do not redistribute raw SEC filing packages or proprietary model outputs as part of this paper. Any released code or derived data will include references to the original resources and follow the applicable terms of use of the source artifacts.

The fields used in our experiments contain company identifiers, filing metadata, audit queries, taxonomy concepts, reporting periods, and numerical ground-truth answers; they do not contain private personal information or user-generated offensive content. Public SEC filings may include publicly disclosed officer or signatory names, but these are not used as prediction targets or evaluation labels.

\section*{Use of AI Assistants}

The authors used AI assistants, including ChatGPT, to support language polishing, brainstorming, and LaTeX editing. All technical claims, experimental results, analysis, and final writing decisions were reviewed and verified by the authors.

\bibliography{custom}

\begin{thebibliography}{38}
\providecommand{\natexlab}[1]{#1}

\bibitem[{Agarwal et~al.(2025)Agarwal, Jomraj, Kaplunov, Krolick, and Rojkova}]{agarwal2025ragulating}
Bhavik Agarwal, Hemant~Sunil Jomraj, Simone Kaplunov, Jack Krolick, and Viktoria Rojkova. 2025.
\newblock Ragulating compliance: A multi-agent knowledge graph for regulatory qa.
\newblock \emph{arXiv preprint arXiv:2508.09893}.

\bibitem[{Amayuelas et~al.(2025)Amayuelas, Sain, Kaur, and Smiley}]{amayuelas2025groundingllmreasoningknowledge}
Alfonso Amayuelas, Joy Sain, Simerjot Kaur, and Charese Smiley. 2025.
\newblock \href {https://arxiv.org/abs/2502.13247} {Grounding llm reasoning with knowledge graphs}.
\newblock \emph{Preprint}, arXiv:2502.13247.

\bibitem[{{Anthropic}(2026{\natexlab{a}})}]{anthropic2026claudecode}
{Anthropic}. 2026{\natexlab{a}}.
\newblock \href {https://code.claude.com/docs/en/overview} {{Claude Code: Overview}}.
\newblock Accessed: 2026-10-10.

\bibitem[{{Anthropic}(2026{\natexlab{b}})}]{anthropic2026claudesonnet46systemcard}
{Anthropic}. 2026{\natexlab{b}}.
\newblock \href {https://www-cdn.anthropic.com/bbd8ef16d70b7a1665f14f306ee88b53f686aa75.pdf} {{System Card: Claude Sonnet 4.6}}.
\newblock Technical report, {Anthropic}.
\newblock Published February 17, 2026. Changelog updated March 6, 2026. Accessed: 2026-05-24.

\bibitem[{Arun et~al.(2025)Arun, Dimino, Agarwal, Sarmah, and Pasquali}]{arun2025finreflectkg}
Abhinav Arun, Fabrizio Dimino, Tejas~Prakash Agarwal, Bhaskarjit Sarmah, and Stefano Pasquali. 2025.
\newblock Finreflectkg: Agentic construction and evaluation of financial knowledge graphs.
\newblock In \emph{Proceedings of the 6th ACM International Conference on AI in Finance}, pages 283--290.

\bibitem[{Berg-Kirkpatrick et~al.(2012)Berg-Kirkpatrick, Burkett, and Klein}]{berg2012empirical}
Taylor Berg-Kirkpatrick, David Burkett, and Dan Klein. 2012.
\newblock An empirical investigation of statistical significance in nlp.
\newblock In \emph{Proceedings of the 2012 joint conference on empirical methods in natural language processing and computational natural language learning}, pages 995--1005.

\bibitem[{Brown et~al.(2020)Brown, Mann, Ryder, Subbiah, Kaplan, Dhariwal, Neelakantan, Shyam, Sastry, Askell et~al.}]{brown2020language}
Tom Brown, Benjamin Mann, Nick Ryder, Melanie Subbiah, Jared~D Kaplan, Prafulla Dhariwal, Arvind Neelakantan, Pranav Shyam, Girish Sastry, Amanda Askell, and 1 others. 2020.
\newblock Language models are few-shot learners.
\newblock \emph{Advances in neural information processing systems}, 33:1877--1901.

\bibitem[{{DeepSeek}(2026)}]{deepseek2026harness}
{DeepSeek}. 2026.
\newblock \href {https://deepseek.com/en/harness/} {{DeepSeek Harness}}.
\newblock Accessed: 2026-10-10.

\bibitem[{Edge et~al.(2024)Edge, Trinh, Cheng, Bradley, Chao, Mody, Truitt, Metropolitansky, Ness, and Larson}]{edge2024local}
Darren Edge, Ha~Trinh, Newman Cheng, Joshua Bradley, Alex Chao, Apurva Mody, Steven Truitt, Dasha Metropolitansky, Robert~Osazuwa Ness, and Jonathan Larson. 2024.
\newblock From local to global: A graph rag approach to query-focused summarization.
\newblock \emph{arXiv preprint arXiv:2404.16130}.

\bibitem[{Fatemi and Hu(2024)}]{fatemi2024enhancing}
Sorouralsadat Fatemi and Yuheng Hu. 2024.
\newblock Enhancing financial question answering with a multi-agent reflection framework.
\newblock In \emph{Proceedings of the 5th ACM International Conference on AI in Finance}, pages 530--537.

\bibitem[{Feng et~al.(2025)Feng, Weir, Bostrom, Bayless, Cassel, Chaudhary, Kiesl-Reiter, and Rangwala}]{feng2025vericotneurosymbolicchainofthoughtvalidation}
Yu~Feng, Nathaniel Weir, Kaj Bostrom, Sam Bayless, Darion Cassel, Sapana Chaudhary, Benjamin Kiesl-Reiter, and Huzefa Rangwala. 2025.
\newblock \href {https://arxiv.org/abs/2511.04662} {Vericot: Neuro-symbolic chain-of-thought validation via logical consistency checks}.
\newblock \emph{Preprint}, arXiv:2511.04662.

\bibitem[{Han et~al.(2024)Han, Kang, Jin, Liu, and Yang}]{han2024xbrl}
Shijie Han, Haoqiang Kang, Bo~Jin, Xiao-Yang Liu, and Steve~Y Yang. 2024.
\newblock Xbrl agent: Leveraging large language models for financial report analysis.
\newblock In \emph{Proceedings of the 5th ACM International Conference on AI in Finance}, pages 856--864.

\bibitem[{Huang et~al.(2026)Huang, Schreyer, Michiles~Jr, and Vasarhelyi}]{huang2026connecting}
Qing Huang, Marco Schreyer, Nilson~Romero Michiles~Jr, and Miklos~A Vasarhelyi. 2026.
\newblock Connecting the dots: Graph neural networks for auditing accounting journal entries.
\newblock \emph{Auditing: A Journal of Practice \& Theory}, pages 1--27.

\bibitem[{Hurst et~al.(2024)Hurst, Lerer, Goucher, Perelman, Ramesh, Clark, Ostrow, Welihinda, Hayes, Radford et~al.}]{hurst2024gpt}
Aaron Hurst, Adam Lerer, Adam~P Goucher, Adam Perelman, Aditya Ramesh, Aidan Clark, AJ~Ostrow, Akila Welihinda, Alan Hayes, Alec Radford, and 1 others. 2024.
\newblock Gpt-4o system card.
\newblock \emph{arXiv preprint arXiv:2410.21276}.

\bibitem[{Lee et~al.(2026)Lee, Kim, Choi, Park, Lyoo, and Park}]{lee2026structureddebateimprovescorporate}
Yoonjin Lee, Munhee Kim, Hanbi Choi, Juhyeon Park, Seungho Lyoo, and Woojin Park. 2026.
\newblock \href {https://arxiv.org/abs/2510.17108} {Structured debate improves corporate credit reasoning in financial ai}.
\newblock \emph{Preprint}, arXiv:2510.17108.

\bibitem[{Lewis et~al.(2020)Lewis, Perez, Piktus, Petroni, Karpukhin, Goyal, K{\"u}ttler, Lewis, Yih, Rockt{\"a}schel et~al.}]{lewis2020retrieval}
Patrick Lewis, Ethan Perez, Aleksandra Piktus, Fabio Petroni, Vladimir Karpukhin, Naman Goyal, Heinrich K{\"u}ttler, Mike Lewis, Wen-tau Yih, Tim Rockt{\"a}schel, and 1 others. 2020.
\newblock Retrieval-augmented generation for knowledge-intensive nlp tasks.
\newblock \emph{Advances in neural information processing systems}, 33:9459--9474.

\bibitem[{Liu et~al.(2026{\natexlab{a}})Liu, Hu, Huang, Niu, Zhang, Ma, Lin, Huat, Kwon, Gao et~al.}]{liu2026evomdt}
Qicai Liu, Zhichao Hu, Tao Huang, Yupeng Niu, Xinche Zhang, Shanwu Ma, Chutong Lin, Goh~Kim Huat, Hyeokkoo~Eric Kwon, Feng Gao, and 1 others. 2026{\natexlab{a}}.
\newblock Evomdt: a self-evolving multi-agent system for structured clinical decision-making in multi-cancer.
\newblock \emph{npj Digital Medicine}.

\bibitem[{Liu et~al.(2026{\natexlab{b}})Liu, Peng, Cao, Wang, Deng, Chen, Yin, and Zhang}]{liu2026toolgatecontractgroundedverifiedtool}
Yanming Liu, Xinyue Peng, Jiannan Cao, Xinyi Wang, Songhang Deng, Jintao Chen, Jianwei Yin, and Xuhong Zhang. 2026{\natexlab{b}}.
\newblock \href {https://arxiv.org/abs/2601.04688} {Toolgate: Contract-grounded and verified tool execution for llms}.
\newblock \emph{Preprint}, arXiv:2601.04688.

\bibitem[{Malarkkan et~al.(2026)Malarkkan, Choudhury, Zhang, Gupta, Wang, Fu, and Zhang}]{malarkkan2026finrulebenchbenchmarkjointreasoning}
Arun~Vignesh Malarkkan, Manan~Roy Choudhury, Guangwei Zhang, Vivek Gupta, Qingyun Wang, Yanjie Fu, and Denghui Zhang. 2026.
\newblock \href {https://arxiv.org/abs/2603.11339} {Finrule-bench: A benchmark for joint reasoning over financial tables and principles}.
\newblock \emph{Preprint}, arXiv:2603.11339.

\bibitem[{{OpenAI}(2026{\natexlab{a}})}]{openai2026gpt61solsystemcard}
{OpenAI}. 2026{\natexlab{a}}.
\newblock \href {https://deploymentsafety.openai.com/gpt-6-1-sol} {{Addendum to GPT-6 Astra System Card: GPT-6.1 Sol}}.
\newblock Accessed: 2026-10-10.

\bibitem[{{OpenAI}(2026{\natexlab{b}})}]{openai2026codexcli}
{OpenAI}. 2026{\natexlab{b}}.
\newblock \href {https://developers.openai.com/codex/cli/} {{Codex CLI}}.
\newblock Accessed: 2026-10-10.

\bibitem[{{OpenClaw}(2026)}]{openclaw2026}
{OpenClaw}. 2026.
\newblock \href {https://github.com/openclaw/openclaw} {{OpenClaw}}.
\newblock Accessed: 2026-10-10.

\bibitem[{Patil et~al.(2024)Patil, Zhang, Wang, and Gonzalez}]{patil2024gorilla}
Shishir~G Patil, Tianjun Zhang, Xin Wang, and Joseph~E Gonzalez. 2024.
\newblock Gorilla: Large language model connected with massive apis.
\newblock \emph{Advances in Neural Information Processing Systems}, 37:126544--126565.

\bibitem[{Peng et~al.(2026)Peng, Xie, Cao, Li, Qian, Wang, Zhang, He, Ai, Ma, Xiang, He, Han, Wang, Guo, Jiang, Zhao, Dong, Wang, Chen, Yuan, Zhang, Lyu, Wu, Yang, Zhao, Dai, Zhang, Elbadry, Gull, Safder, Chen, Zhu, Cai, Wang, Giannouris, Jiang, Liu, Kabir, Wang, Zheng, Yu, Liu, Cao, Xu, Lu, Huang, Mo, Lin, Tiwari, Zhao, Basulto, Liu, Smith, Pei, Cohan, Huang, Tang, Lopez-Lira, Chen, Liu, Tsujii, Nie, and Ananiadou}]{peng2026herculeanagenticbenchmarkfinancial}
Xueqing Peng, Zhuohan Xie, Yupeng Cao, Haohang Li, Lingfei Qian, Yan Wang, Vincent~Jim Zhang, Huan He, Xuguang Ai, Linhai Ma, Ruoyu Xiang, Yueru He, Yi~Han, Shuyao Wang, Yuqing Guo, Mingyang Jiang, Yilun Zhao, Youzhong Dong, Xiaoyu Wang, and 45 others. 2026.
\newblock \href {https://arxiv.org/abs/2605.14355} {Herculean: An agentic benchmark for financial intelligence}.
\newblock \emph{Preprint}, arXiv:2605.14355.

\bibitem[{Qian et~al.(2025)Qian, Zhou, Wang, Peng, Huang, and Xie}]{qian2025fino1}
Lingfei Qian, Weipeng Zhou, Yan Wang, Xueqing Peng, Jimin Huang, and Qianqian Xie. 2025.
\newblock Fino1: On the transferability of reasoning enhanced llms to finance.
\newblock \emph{arXiv preprint arXiv:2502.08127}.

\bibitem[{{Qwen Team}(2026{\natexlab{a}})}]{qwen3.5}
{Qwen Team}. 2026{\natexlab{a}}.
\newblock \href {https://qwen.ai/blog?id=qwen3.5} {{Qwen3.5}: Towards native multimodal agents}.

\bibitem[{{Qwen Team}(2026{\natexlab{b}})}]{qwen3.6-27b}
{Qwen Team}. 2026{\natexlab{b}}.
\newblock \href {https://qwen.ai/blog?id=qwen3.6-27b} {{Qwen3.6-27B}: Flagship-level coding in a {27B} dense model}.

\bibitem[{Sarthi et~al.(2024)Sarthi, Abdullah, Tuli, Khanna, Goldie, and Manning}]{sarthi2024raptor}
Parth Sarthi, Salman Abdullah, Aditi Tuli, Shubh Khanna, Anna Goldie, and Christopher Manning. 2024.
\newblock Raptor: Recursive abstractive processing for tree-organized retrieval.
\newblock In \emph{International Conference on Learning Representations}, volume 2024, pages 32628--32649.

\bibitem[{Schick et~al.(2023)Schick, Dwivedi-Yu, Dess{\`\i}, Raileanu, Lomeli, Hambro, Zettlemoyer, Cancedda, and Scialom}]{schick2023toolformer}
Timo Schick, Jane Dwivedi-Yu, Roberto Dess{\`\i}, Roberta Raileanu, Maria Lomeli, Eric Hambro, Luke Zettlemoyer, Nicola Cancedda, and Thomas Scialom. 2023.
\newblock Toolformer: Language models can teach themselves to use tools.
\newblock \emph{Advances in neural information processing systems}, 36:68539--68551.

\bibitem[{Wang et~al.(2025)Wang, Chi, Tai, Kwok, He, Li, Hua, Li, Lu, Wang et~al.}]{wang2025finsage}
Xinyu Wang, Jijun Chi, Zhenghan Tai, Tung Sum~Thomas Kwok, Hailin He, Zhuhong Li, Yuchen Hua, Muzhi Li, Peng Lu, Suyucheng Wang, and 1 others. 2025.
\newblock Finsage: A multi-aspect rag system for financial filings question answering.
\newblock In \emph{Proceedings of the 34th ACM International Conference on Information and Knowledge Management}, pages 6144--6152.

\bibitem[{Wang et~al.(2026)Wang, Wang, Yang, Patel, Zhao, Mo, Peng, Qian, Chen, Guti\'{e}rrez-Basulto, Huang, Xiong, Liu, and Nie}]{10.1145/3805712.3808578}
Yan Wang, Keyi Wang, Shanshan Yang, Jaisal Patel, Jeff Zhao, Fengran Mo, Xueqing Peng, Lingfei Qian, Yankai Chen, V\'{\i}ctor Guti\'{e}rrez-Basulto, Jimin Huang, Guojun Xiong, Xiao-Yang Liu, and Jian-Yun Nie. 2026.
\newblock \href {https://doi.org/10.1145/3805712.3808578} {Finauditing: A financial taxonomy-structured multi-document benchmark for evaluating llms}.
\newblock In \emph{Proceedings of the 49th International ACM SIGIR Conference on Research and Development in Information Retrieval}, SIGIR '26, pages 3456--3463, New York, NY, USA. Association for Computing Machinery.

\bibitem[{Yang and Xu(2013)}]{yang2013evidential}
Jian-Bo Yang and Dong-Ling Xu. 2013.
\newblock Evidential reasoning rule for evidence combination.
\newblock \emph{Artificial Intelligence}, 205:1--29.

\bibitem[{Yang et~al.(2026)Yang, Li, Qiang, Wang, Lou, Li, Cheng, Xu, Lian, Zhang et~al.}]{yang2026finvault}
Zhi Yang, Runguo Li, Qiqi Qiang, Jiashun Wang, Fangqi Lou, Mengping Li, Dongpo Cheng, Rui Xu, Heng Lian, Shuo Zhang, and 1 others. 2026.
\newblock Finvault: Benchmarking financial agent safety in execution-grounded environments.
\newblock \emph{arXiv preprint arXiv:2601.07853}.

\bibitem[{Yao et~al.(2023)Yao, Zhao, Yu, Du, Shafran, Narasimhan, and Cao}]{yao2023reactsynergizingreasoningacting}
Shunyu Yao, Jeffrey Zhao, Dian Yu, Nan Du, Izhak Shafran, Karthik Narasimhan, and Yuan Cao. 2023.
\newblock \href {https://arxiv.org/abs/2210.03629} {React: Synergizing reasoning and acting in language models}.
\newblock \emph{Preprint}, arXiv:2210.03629.

\bibitem[{Yin et~al.(2026)Yin, Sha, Cui, Meng, and Li}]{yin2026reasoningtrapenhancingllm}
Chenlong Yin, Zeyang Sha, Shiwen Cui, Changhua Meng, and Zechao Li. 2026.
\newblock \href {https://arxiv.org/abs/2510.22977} {The reasoning trap: How enhancing llm reasoning amplifies tool hallucination}.
\newblock \emph{Preprint}, arXiv:2510.22977.

\bibitem[{Zhang et~al.(2026{\natexlab{a}})Zhang, Song, Elbadry, Chen, Wang, Zhou, Zheng, He, Dai, Georgiev, Gull, Safder, Wu, Meng, Ji, Zhao, Peng, Huang, Chen, Xue, Liu, Nakov, and Xie}]{zhang2026finreportingagenticworkflowlocalized}
Fan Zhang, Mingzi Song, Rania Elbadry, Yankai Chen, Shaobo Wang, Yixi Zhou, Xunwen Zheng, Yueru He, Yuyang Dai, Georgi Georgiev, Ayesha Gull, Muhammad~Usman Safder, Fan Wu, Liyuan Meng, Fengxian Ji, Junning Zhao, Xueqing Peng, Jimin Huang, Yu~Chen, and 4 others. 2026{\natexlab{a}}.
\newblock \href {https://arxiv.org/abs/2604.05966} {Finreporting: An agentic workflow for localized reporting of cross-jurisdiction financial disclosures}.
\newblock \emph{Preprint}, arXiv:2604.05966.

\bibitem[{Zhang et~al.(2026{\natexlab{b}})Zhang, Seedat, Dong, Cui, Zhu, and van~de Schaar}]{zhang2026guidelinegroundedevidenceaccumulationhighstakes}
Yichi Zhang, Nabeel Seedat, Yinpeng Dong, Peng Cui, Jun Zhu, and Mihaela van~de Schaar. 2026{\natexlab{b}}.
\newblock \href {https://arxiv.org/abs/2603.02798} {Guideline-grounded evidence accumulation for high-stakes agent verification}.
\newblock \emph{Preprint}, arXiv:2603.02798.

\bibitem[{Zhong et~al.(2024)Zhong, Yang, Shi, Wei, and Wang}]{zhong2024data}
Hao Zhong, Dong Yang, Shengdong Shi, Lai Wei, and Yanyan Wang. 2024.
\newblock From data to insights: the application and challenges of knowledge graphs in intelligent audit.
\newblock \emph{Journal of Cloud Computing}, 13(1):114.

\end{thebibliography}

\clearpage
\appendix
\section{Extended Related Work}
\label{app:related-work}

\paragraph{Grounded and agentic verification.}
Prior work improves LLM reliability through tool use, retrieval, graph
grounding, neuro-symbolic validation, and multi-agent
interaction~\citep{yao2023reactsynergizingreasoningacting,
schick2023toolformer, patil2024gorilla, edge2024local,
amayuelas2025groundingllmreasoningknowledge,
feng2025vericotneurosymbolicchainofthoughtvalidation,
fatemi2024enhancing, lee2026structureddebateimprovescorporate,
agarwal2025ragulating}. These methods expand evidence access and support cross-checking. Tool-output interpretation and validation remain active research questions~\citep{liu2026toolgatecontractgroundedverifiedtool,yin2026reasoningtrapenhancingllm}; their guarantees depend on the tools and constraints of each system. In \textsc{AuditFlow}, typed deterministic
operators execute the numerical checks and rule evaluation, while agents guide
the evidence search. A required-tool gate constrains ordinary junior termination; budget exhaustion follows a separate fallback path.

\paragraph{Clinical and legal verification.}
Clinical guideline checking and legal compliance similarly require decisions
grounded in external rules and supporting evidence~\citep{
zhang2026guidelinegroundedevidenceaccumulationhighstakes,
liu2026evomdt}. XBRL numerical consistency verification further requires exact
recomputation over machine-readable facts governed by filing-specific
accounting structures. This includes calculation, dimensional, period, and unit
constraints, which limits the direct transfer of methods centered primarily on
textual rule alignment.

\paragraph{Financial language models and structured auditing.}
Recent financial systems use retrieval, tools, graphs, and agents to analyze
filings and disclosures~\citep{han2024xbrl, wang2025finsage,
arun2025finreflectkg, zhang2026finreportingagenticworkflowlocalized,
peng2026herculeanagenticbenchmarkfinancial}. Related audit research uses
knowledge graphs and graph neural networks to represent accounting structure
and detect suspicious patterns~\citep{zhong2024data, huang2026connecting}.
These studies demonstrate the value of structured financial evidence.
\textsc{AuditFlow} focuses on query-specific verification by linking reusable
US-GAAP constraints with a graph constructed from each filing and executing
fact retrieval, relation traversal, expected-value reconstruction, and
rule-level checks within the same environment.

\paragraph{Financial verification benchmarks and positioning.}
FinAuditing introduces the FinMR numerical-consistency task used in this
work~\citep{10.1145/3805712.3808578}, while FinRule-Bench and FinVault evaluate
financial rule reasoning and safe agent execution~\citep{
malarkkan2026finrulebenchbenchmarkjointreasoning, yang2026finvault}. These
benchmarks show that reliable financial verification requires more than
evidence retrieval. \textsc{AuditFlow} organizes graphs, tools, and
role-specialized agents around \emph{search-computation separation}: agents
adaptively explore filing and taxonomy evidence, while the executable symbolic environment produces numerical and rule-specific evidence. Evidential
aggregation then combines the structured reports without using self-reported
LLM confidence.

\section{Tool Access and Tool Inventory}
\label{app:tools}

Table~\ref{tab:agent-tools} lists role-specific access; Table~\ref{tab:tool-inventory} describes each tool.

\paragraph{Agent-specific tool access.}
The Compliance Auditor $A_1$ receives the static catalog: concept metadata, calculation trees, dimensional axes, rule lookup, fact retrieval, and deterministic checkers. The Forensic Auditor $A_2$ additionally receives dynamic tools for historical, dimensional, sibling, unit, and magnitude analyses.

Both catalogs access filing evidence. Their distinction specifies available investigation paths rather than separating access to the two graphs.

The Senior Auditor $A_s$ has no tool access in the reported experiments. A programmatic check accepts agreement on the junior verdicts and rule sets; otherwise, the senior provides feedback for refinement. A tool-enabled senior exists in the implementation but is not evaluated here.

\paragraph{Required-tool constraints.}
The two junior auditors are allowed to choose their own search trajectories, but their ordinary finalization and senior-request actions require non-error observations from the mandatory tools. For $A_1$, the required tools are \texttt{check\_sign}, \texttt{check\_calc\_tree}, and \texttt{check\_\allowbreak dim\_\allowbreak consistency}. For $A_2$, we additionally require \texttt{get\_fact\_history}, which ensures that the evidence-first path is exercised before the report is accepted. If a junior auditor attempts to stop early, the environment rejects the terminal action and returns the missing tools as the next observation. The gate tests execution, not whether every rule was conclusively evaluated. Forced output after budget exhaustion bypasses this gate and does not run missing checks (Appendix~\ref{app:implementation}).

\begin{table*}[t]
\centering
\small
\setlength{\tabcolsep}{4pt}
\begin{tabular}{p{0.13\linewidth} p{0.18\linewidth} p{0.43\linewidth} p{0.18\linewidth}}
\toprule
Agent & Catalog & Tools & Required for ordinary termination \\
\midrule
$A_1$ Compliance &
Static taxonomy &
\texttt{lookup\_concept\_spec};
\texttt{walk\_calc\_tree};
\texttt{list\_dim\_axes};
\texttt{find\_dqc\_rules};
\texttt{check\_sign};
\texttt{check\_calc\_tree};
\texttt{check\_\allowbreak dim\_\allowbreak consistency};
\texttt{get\_fact} &
\texttt{check\_sign};
\texttt{check\_calc\_tree};
\texttt{check\_\allowbreak dim\_\allowbreak consistency} \\
\midrule
$A_2$ Forensic &
Static + dynamic &
All $A_1$ tools, plus
\texttt{get\_fact\_history};
\texttt{compare\_periods};
\texttt{detect\_temporal\_outlier};
\texttt{get\_dimensional\_breakdown};
\texttt{compare\_dim\_members};
\texttt{check\_sibling\_math\_consistency};
\texttt{check\_magnitude\_plausibility};
\texttt{check\_unit\_decimal\_consistency};
\texttt{find\_related\_concepts} &
\texttt{get\_fact\_history};
\texttt{check\_sign};
\texttt{check\_calc\_tree};
\texttt{check\_\allowbreak dim\_\allowbreak consistency} \\
\midrule
$A_s$ Senior &
None in the reported experiments &
Emits refinement feedback when junior reports disagree. &
Not applicable \\
\bottomrule
\end{tabular}
\caption{Agent-specific tool access in \textsc{AuditFlow}. The two junior auditors use the same ReAct engine but different tool catalogs and terminal constraints. Required tools must return non-error observations before ordinary finalization or a senior request; budget-exhaustion fallback is separate.}
\label{tab:agent-tools}
\end{table*}

\paragraph{Tool inventory.}
Table~\ref{tab:tool-inventory} lists all tools exposed by the audit environment. We group them into three categories. Static taxonomy tools expose the regulatory subspace, including concept definitions, calculation relationships, dimensional axes, and candidate DQC rules. Static check tools execute the terminal rule-specific verification steps for the three DQC rule families evaluated in this work. Dynamic filing tools expose filing-specific evidence used mainly by the Forensic Auditor, including historical facts, period comparisons, dimensional breakdowns, sibling relationships, and unit or decimal consistency.

Static and dynamic refer to tool-catalog categories, not exclusive access to one graph. The static catalog includes taxonomy queries, deterministic checks, and \texttt{get\_fact}; the latter two also read filing facts. Dynamic tools provide additional filing analyses. Tool-use shares follow this catalog classification. The deterministic checkers connect both graphs by applying rule constraints to reported facts.

\begin{table*}[t]
\centering
\small
\setlength{\tabcolsep}{4pt}
\begin{tabular}{@{}p{0.18\linewidth} p{0.30\linewidth} p{0.48\linewidth}@{}}
\toprule
Tool group & Tool & Observation returned to the agent \\
\midrule
Static taxonomy &
\texttt{lookup\_concept\_spec} &
Concept metadata: balance, period type, data type, label, definition, and abstract flag. \\
Static taxonomy &
\texttt{walk\_calc\_tree} &
Weighted calculation-linkbase neighborhood, upward or downward, around the target concept. \\
Static taxonomy &
\texttt{list\_dim\_axes} &
Dimensional axes associated with the target concept. \\
Static taxonomy &
\texttt{find\_dqc\_rules} &
Conservative list of potentially applicable DQC rule families inferred from taxonomy structure. \\
Static check &
\texttt{check\_sign} &
DQC.US.0015 verdict over non-dimensional and dimensional facts, including the offending value and balance attribute when a violation is found. \\
Static check &
\texttt{check\_calc\_tree} &
DQC.US.0126 verdict, expected value $v_{\mathrm{exp}}$, and weighted child-sum breakdown. \\
Static check &
\texttt{check\_\allowbreak dim\_\allowbreak consistency} &
DQC.US.0117 verdict, non-dimensional total, candidate dimensional sums, dimensional facts, and ambiguity flag. \\
Static taxonomy &
\texttt{get\_fact} &
All reported facts for a concept and period, including value, unit, decimals, period, and dimensional context. \\
\midrule
Dynamic filing &
\texttt{get\_fact\_history} &
Filing-internal comparative periods for the same concept, or a no-history signal. \\
Dynamic filing &
\texttt{compare\_periods} &
Delta and percentage change between two reported periods of the same concept. \\
Dynamic filing &
\texttt{detect\_temporal\_outlier} &
Temporal anomaly signal against filing-internal history. \\
Dynamic filing &
\texttt{get\_dimensional\_breakdown} &
Dimensional facts grouped by unit and axis tuple. \\
Dynamic filing &
\texttt{compare\_dim\_members} &
Member shares, dimensional sum, and match against the non-dimensional total. \\
Dynamic filing &
\texttt{check\_sibling\_math\_consistency} &
Sibling arithmetic around parents that contain the target concept in the filing calculation tree. \\
Dynamic filing &
\texttt{check\_magnitude\_plausibility} &
Order-of-magnitude anomaly signal relative to other non-dimensional facts in the same filing. \\
Dynamic filing &
\texttt{check\_unit\_decimal\_consistency} &
Unit and decimals consistency checks for all target facts. \\
Dynamic filing &
\texttt{find\_related\_concepts} &
Parents, children, and siblings structurally adjacent to the target concept inside the filing graph. \\
\bottomrule
\end{tabular}
\caption{Complete tool inventory. Static tools expose the regulatory subspace and deterministic DQC checks; dynamic tools expose filing-specific evidence used by the Forensic Auditor.}
\label{tab:tool-inventory}
\end{table*}

\paragraph{Role of deterministic checkers.}
The three required DQC checkers are the terminal verification tools used by the system. They are not ordinary retrieval operations. Each checker returns a structured rule outcome and the audit-critical values needed by the evaluator, including the reported value, expected value when applicable, supporting facts, and rule-specific findings. Numerical operations remain deterministic in both coverage conditions. The authoritative configuration follows the checker outcomes; the override-enabled configuration additionally allows an evidence-supported violation to be reported when a checker is silent. The prompt requires such overrides to be identified and justified, and prohibits discarding positive checker findings. This permission is varied in Appendix~\ref{app:coverage}.

\section{Graph Construction and Deterministic Tool Execution}
\label{app:graph-construction}

This section describes how the dual graph is constructed and how deterministic tools operate over it. Both the static taxonomy graph and the dynamic filing graph are represented as \texttt{networkx.DiGraph} objects. Their construction contains no learned component and does not involve an LLM. The static graph is built once for each US-GAAP taxonomy release and reused across filings from the same reporting year. The dynamic graph is constructed separately for each XBRL filing package. Both graphs are cached after construction.

\paragraph{Graph schema.}
Table~\ref{tab:graph-schema} summarizes the node types, edge types, and attributes used in the dual-graph environment. The taxonomy graph stores regulatory concepts and relationships. The filing graph stores reported facts and their filing-specific contexts. Bridge edges connect each reported fact to the taxonomy or extension concept that governs it.

\begin{table*}[t]
\centering
\small
\setlength{\tabcolsep}{4pt}
\begin{tabular}{p{0.13\linewidth} p{0.18\linewidth} p{0.25\linewidth} p{0.36\linewidth}}
\toprule
Graph & Element & Type & Key attributes \\
\midrule
Taxonomy &
Node &
Concept &
Concept identifier, label, definition, balance, period type, data type, and abstract flag. \\

Taxonomy &
Edge &
Calculation &
Parent concept, child concept, arcrole, order, and calculation weight. \\

Taxonomy &
Edge &
Presentation &
Parent concept, child concept, arcrole, and presentation order. \\

Taxonomy &
Edge &
Dimensional &
Hypercube, axis, domain, member, arcrole, and target role. \\
\midrule

Filing &
Node &
Fact &
Concept identifier, reported value, unit, decimals, context identifier, period, and dimensional assignments. \\

Filing &
Node &
Context &
Entity identifier, start date, end date or instant, and dimensional assignments. \\

Filing &
Node &
Unit &
Unit identifier and numerical measure. \\

Filing &
Node &
Extension concept &
Namespace, label, balance, period type, and data type when available. \\

Filing &
Edge &
Evidence relation &
Fact-to-context, fact-to-unit, context-to-period, and context-to-dimension relations. \\

Filing &
Edge &
Filing-specific relation &
Calculation, presentation, or dimensional relation defined by the filing linkbases. \\
\midrule

Bridge &
Edge &
Fact-to-concept &
Link from a reported fact to its governing US-GAAP or extension concept. \\
\bottomrule
\end{tabular}
\caption{Node and edge schema of the dual-graph environment. The taxonomy graph stores reusable regulatory structure, while the filing graph stores filing-specific facts and contexts.}
\label{tab:graph-schema}
\end{table*}

\paragraph{Deterministic graph construction.}
Algorithm~\ref{alg:graph-construction} summarizes the construction procedure. We first parse the corresponding US-GAAP taxonomy release and create one taxonomy node for each concept. Calculation, presentation, and dimensional linkbases are converted into typed directed edges. We then parse the filing instance document and create nodes for facts, contexts, units, periods, dimensions, and extension concepts. Filing-specific linkbases are added to the filing graph, and bridge edges connect facts to the concepts that define them.

The XML syntax and linkbase formats are standardized, so basic element parsing is largely mechanical. Constructing an audit-ready graph nevertheless requires additional normalization and linking. Concept identifiers are canonicalized using namespace URIs rather than document-specific prefixes. Contexts are normalized by entity, period, and dimensional assignments because different context identifiers may encode the same logical context. Units are resolved before facts are grouped for numerical comparison. Filing-specific extension concepts are retained rather than discarded. We also preserve extended link
roles, since the same concept may occur in multiple calculation or presentation networks. These steps prevent raw XML identifiers from being treated as semantic identities.

\begin{algorithm}[t]
\small
\caption{Deterministic Dual-Graph Construction}
\label{alg:graph-construction}
\begin{algorithmic}[1]
\Require US-GAAP taxonomy release $\mathcal{T}_y$, XBRL filing package $\mathcal{F}$
\Ensure Taxonomy graph $\mathcal{G}^{y}_{T}$, filing graph $\mathcal{G}_{F}$

\State Initialize directed graphs $\mathcal{G}^{y}_{T}$ and $\mathcal{G}_{F}$
\State Canonicalize concept identifiers by namespace URI and local name

\ForAll{concepts and relations in $\mathcal{T}_y$}
    \State Add concept nodes with taxonomy metadata
    \State Add typed calculation, presentation, and dimensional edges
    \State Preserve extended link roles and relation-specific attributes
\EndFor

\ForAll{contexts, units, periods, dimensions, and extension concepts in $\mathcal{F}$}
    \State Normalize contexts and units
    \State Add the corresponding nodes and typed relations to $\mathcal{G}_{F}$
\EndFor

\ForAll{reported facts $f$ in $\mathcal{F}$}
    \State Add $f$ with its value, decimals, period, unit, context, and dimensions
    \State Link $f$ to its normalized context, unit, period, dimensions, and governing concept
\EndFor

\ForAll{filing-specific calculation, presentation, and dimensional relations}
    \State Add typed edges with extended link roles and relation attributes
\EndFor

\State Cache and return $\mathcal{G}^{y}_{T}$ and $\mathcal{G}_{F}$
\end{algorithmic}
\end{algorithm}

\paragraph{Deterministic tool execution.}
Each tool is implemented as a deterministic Python function over the constructed graphs. The LLM selects a tool and supplies structured arguments, but it does not generate the values returned by the tool. Given the same graph state and arguments, the tool returns the same reported value, expected value, rule flags, and supporting evidence regardless of the LLM backbone. Tool outputs preserve provenance by including the facts, contexts, relationships, and intermediate values used in each decision.

Algorithm~\ref{alg:calc-check} summarizes \texttt{check\_calc\_tree}. At each visited concept, filing-specific calculation children take precedence over taxonomy children. Fact lookup selects the first non-dimensional fact matching the concept and queried period; it does not add a unit-equality or context-identifier check. The checker skips a level if its parent or any direct child fact is missing, and continues into descendants. It therefore does not report a violation from a partial child sum. The traversal uses a visited set and an optional depth limit; the reported top-level expected value is computed separately and is null if its child set is empty or incomplete. The absolute comparison tolerance is the implementation default $\epsilon=0.01$; the tool wrapper does not derive it from XBRL decimals.

\begin{algorithm}[t]
\small
\caption{Deterministic Calculation-Tree Checker}
\label{alg:calc-check}
\begin{algorithmic}[1]
\Require Concept $c$, period $t$, graphs $\mathcal{G}^{y}_{T}$ and $\mathcal{G}_{F}$, tolerance $\epsilon=0.01$
\Ensure Checker verdict, top-level expected value, and level violations
\State $\mathcal{V}\gets\emptyset$
\ForAll{concepts $u$ visited from $c$ within the depth limit}
    \State $\mathcal{C}_u\gets$ filing calculation children, or taxonomy children if absent
    \State Retrieve first non-dimensional parent and child facts matching $t$
    \If{$\mathcal{C}_u\ne\emptyset$ and parent and all child facts exist}
        \State $s_u\gets\sum_{(u_i,w_i)\in\mathcal{C}_u}w_i v_i$
        \If{$|v_u-s_u|>\epsilon$}
            \State Append $(u,v_u,s_u,\{(u_i,w_i,v_i)\})$ to $\mathcal{V}$
        \EndIf
    \Else
        \State Skip this level's comparison
    \EndIf
    \State Continue to unvisited children within the depth limit
\EndFor
\State $\mathcal{C}_c\gets$ filing calculation children of $c$, or taxonomy children if absent
\If{$\mathcal{C}_c\ne\emptyset$ and all child facts matching $t$ exist}
    \State $v_{\mathrm{exp}}\gets\sum_{(c_i,w_i)\in\mathcal{C}_c}w_i v_i$
\Else
    \State $v_{\mathrm{exp}}\gets\texttt{null}$
\EndIf
\State $\mathrm{verdict}\gets\texttt{violation}$ if $\mathcal{V}\ne\emptyset$, else \texttt{ok}
\State \Return $\mathrm{verdict}$, $v_{\mathrm{exp}}$, and $\mathcal{V}$
\end{algorithmic}
\end{algorithm}

The internal calculation result records skipped levels in its trace. The tool wrapper exposes the top-level expected value and any level violations with their weighted breakdowns, but does not return those skip messages or a distinct \texttt{not\_evaluable} status. Its \texttt{ok} outcome means that no violation was found among evaluated levels, not that every applicable relationship was resolved. This distinction also limits what successful execution of the required-tool gate establishes.

The other checkers return rule-specific payloads. \texttt{check\_sign} returns \texttt{no\_data} when no matching facts exist, or a sign outcome with details of any violating fact. \texttt{check\_\allowbreak dim\_\allowbreak consistency} returns the non-dimensional total, candidate dimensional sums, and an ambiguity flag when a violation leaves multiple candidate groups unresolved. These outputs provide deterministic evidence, but do not implement the full official rule engine.

\subsection{Evidential Aggregation Details}
\label{app:fusion}

\textsc{AuditFlow} derives the final verdict from structured evidence rather than from an LLM confidence score. Each junior report records the triggered rules, reported value, expected value when available, terminal status, and supporting tool trace. The mass function does not use the free-form rationale or the model's self-reported confidence.


Each junior report defines a mass function
$m_i:\{\mathrm{v},\mathrm{c},\Theta\}\rightarrow[0,1]$ for
$i\in\{1,2\}$, where $\mathrm{v}$, $\mathrm{c}$, and $\Theta$ denote
\textsc{violation}, \textsc{consistent}, and unresolved evidence, respectively,
with
\begin{equation}
m_i(\mathrm{v})+m_i(\mathrm{c})+m_i(\Theta)=1.
\end{equation}
Table~\ref{tab:mass-mapping} gives the deterministic mapping.

\begin{table}[t]
\centering
\small
\setlength{\tabcolsep}{3pt}
\resizebox{\columnwidth}{!}{
\begin{tabular}{lccc}
\toprule
Junior report state &
$m(\mathrm{v})$ &
$m(\mathrm{c})$ &
$m(\Theta)$ \\
\midrule
DQC.US.0015 reported & 0.95 & 0.00 & 0.05 \\
DQC.US.0126 reported & 0.95 & 0.00 & 0.05 \\
DQC.US.0117 reported & 0.70 & 0.00 & 0.30 \\
No rule; value resolved & 0.00 & 0.85 & 0.15 \\
No rule; value unresolved & 0.00 & 0.40 & 0.60 \\
\bottomrule
\end{tabular}
}
\caption{Deterministic mapping from a structured junior report to a mass function.}
\label{tab:mass-mapping}
\end{table}

Violation mass is assigned from the rules in the structured junior report. Under authoritative checking, these follow positive checker observations; with overrides enabled, they may also include violations the agent identifies from the evidence despite a silent check. If several rules are reported, we combine their fixed rule-specific reliabilities as independent simple support functions; these weights are not estimated from the evaluation set. For triggered rules with reliabilities $\rho_1,\ldots,\rho_k$, the resulting mass is
\begin{equation}
m(\mathrm{v}) = 1-\prod_{j=1}^{k}(1-\rho_j), \qquad
m(\Theta) = \prod_{j=1}^{k}(1-\rho_j),
\end{equation}
with $m(\mathrm{c})=0$. For example, simultaneous DQC.US.0015 and DQC.US.0126 violations produce
$m(\mathrm{v})=0.9975$ and $m(\Theta)=0.0025$.

When no rule is reported, the mass-mapping implementation treats evidence as complete when the reported value has been resolved. The gate separately checks required-tool execution on ordinary termination paths. If the reported value is unresolved, more mass is assigned to $\Theta$. Thus, the completeness flag is an operational indicator rather than a guarantee that all relevant evidence has been found.


Let $m_1$ and $m_2$ be the mass functions produced by $A_1$ and $A_2$.
We combine them using evidential reasoning~\citep{yang2013evidential}.
For each decision state $X\in \{\mathrm{v},\mathrm{c}\}$,
\begin{align}
  \tilde{m}(X)
  &= m_1(X)m_2(X)
   + m_1(X)m_2(\Theta) \notag\\
  &\quad + m_1(\Theta)m_2(X), \\
  \tilde{m}(\Theta)
  &= m_1(\Theta)m_2(\Theta), \\
  K
  &= m_1(\mathrm{v})m_2(\mathrm{c})
   + m_1(\mathrm{c})m_2(\mathrm{v}),
\end{align}
where $K$ is the conflict mass. The normalized fused mass ($m^+$ in Figure~\ref{fig:overview}) is
\begin{equation}
  m^\star(\cdot)=\frac{\tilde{m}(\cdot)}{1-K}.
\end{equation}
We define the trustworthiness score as
\begin{equation}
  \tau = 1 - m^\star(\Theta).
\end{equation}

For the final binary decision, the remaining ignorance mass is redistributed over \textsc{violation} and \textsc{consistent}. The system predicts \textsc{violation} when the redistributed violation mass is larger. This aggregation exposes two signals: $K$ measures disagreement between the junior auditors, and $\tau$ measures the mass assigned to the two decision states after fusion, not a calibrated probability of correctness.

\paragraph{Numerical output selection.}
Agreement compares junior verdicts and reported rule sets, not numerical fields. After fusion, the pipeline selects a junior whose verdict matches the fused decision, preferring $A_1$ if both match; if neither matches, it selects the report with more rules, breaking ties in favor of $A_1$. The final expected value, rules, and supporting evidence come from this report; the reported value comes from the shared audit state. Numerical fields are not averaged or checked for agreement by fusion. The output translator then selects the numerical candidate associated with the reported rules (Appendix~\ref{app:metrics}).

\subsection{Complete Audit Procedure}
\label{app:audit-procedure}

\begin{algorithm}[t]
\small
\caption{Environment-grounded Multi-Agent Auditing}
\label{alg:audit}
\begin{algorithmic}[1]
\Require Audit environment
$\mathcal{E}=(\mathcal{G}^{y}_{T},\mathcal{G}_{F},\mathcal{A},O)$,
query $(c,t)$, maximum refinement rounds $R_{\max}$
\Ensure $\hat{y}$, $v_{\mathrm{rep}}$, $v_{\mathrm{exp}}$,
$\mathcal{R}$, $\mathcal{P}$, $\mathcal{Z}$, $\tau$
\State Initialize shared audit state from query $(c,t)$
\State Retrieve reported value $v_{\mathrm{rep}}$ from $\mathcal{G}_{F}$
\For{$k=0$ \textbf{to} $R_{\max}$}
  \State Run Compliance Auditor $A_1$ and Forensic Auditor $A_2$ over
  $\mathcal{E}$
  \State Gate ordinary terminal actions; on budget exhaustion use available observations
  \If{$A_1$ and $A_2$ agree on decision and triggered rules}
     \State \Return $\textsc{Aggregate}(A_1,A_2)$
  \EndIf
  \If{$k<R_{\max}$}
    \State $\textit{feedback}\gets A_s(A_1,A_2)$
    \State Append $\textit{feedback}$ to the shared audit state
  \EndIf
\EndFor
\State \Return $\textsc{Aggregate}(A_1,A_2)$
\end{algorithmic}
\end{algorithm}

\paragraph{Mass-function assignment.}
Algorithm~\ref{alg:mass-mapping} shows how a structured junior report is converted into a mass function. The mapping reads the structured report's triggered rules and whether its reported value has been resolved. Override-enabled reports can include agent-identified violations in addition to positive checker findings. The mass function itself does not independently validate an override or use the free-form confidence score.

\begin{algorithm}[t]
\small
\caption{Deterministic Mass-Function Assignment}
\label{alg:mass-mapping}
\begin{algorithmic}[1]
\Require Structured junior report $R$
\Ensure Mass function $m$ over $\{\mathrm{v},\mathrm{c},\Theta\}$

\State $\mathcal{R}\gets$ distinct triggered rules reported in $R$
\State $\mathrm{complete}\gets v_{\mathrm{rep}}$ is resolved

\If{$\mathcal{R}$ is not empty}
\State Assign $\rho_r=0.95$ for
$r\in\{\texttt{DQC.US.0015},\texttt{DQC.US.0126}\}$
\State Assign $\rho_r=0.70$ for $r=\texttt{DQC.US.0117}$
\State $m(\Theta)\gets\prod_{r\in\mathcal{R}}(1-\rho_r)$
\State $m(\mathrm{v})\gets 1-m(\Theta)$
\State $m(\mathrm{c})\gets 0$
\ElsIf{$\mathrm{complete}$}
\State $(m(\mathrm{v}),m(\mathrm{c}),m(\Theta))
\gets(0.00,0.85,0.15)$
\Else
\State $(m(\mathrm{v}),m(\mathrm{c}),m(\Theta))
\gets(0.00,0.40,0.60)$
\EndIf

\State \Return $m$
\end{algorithmic}
\end{algorithm}

\section{Dataset Details}
\label{app:dataset}

\paragraph{Data sources and instance identity.}
The 1{,}000-instance evaluation set was collected in two rounds following FinAuditing's FinMR methodology. The rounds share no instance with the same ticker, filing type, filing release date, target concept, and reporting period. Comparing reference values numerically rather than as strings also yields no duplicate rows. Each instance retains its source-round and original identifiers for traceability. Filing dates span 2021--2026; queried periods also include comparative facts from 2019--2020. We construct US-GAAP taxonomy graphs for all six release years from 2021 to 2026 and use the release corresponding to each filing.

\paragraph{Evaluation set and subsets.}
Table~\ref{tab:dataset} summarizes the evaluation set and the two subsets used in the experiments. An instance is a concept--period audit case; a single filing or issuer can therefore contribute multiple instances. Filing counts refer to distinct combinations of ticker, filing type, and release date, while issuer counts refer to distinct tickers. The evaluation set contains 578 sign-consistency cases, 71 dimensional-aggregation cases, and 351 calculation-tree cases. The 300-instance ablation subset contains 174, 21, and 105 cases of these rules, respectively; the 100-instance hard subset contains 58, 7, and 35. Both subsets are drawn from the evaluation set and approximately preserve its rule proportions.

\begin{table}[t]
\centering
\small
\setlength{\tabcolsep}{4pt}
\begin{tabular}{lrrr}
\toprule
 & Evaluation & Ablation & Hard \\
\midrule
Instances & 1{,}000 & 300 & 100 \\
Filings & 548 & 236 & 69 \\
Issuers & 319 & 167 & 44 \\
\midrule
DQC.US.0015 & 57.8\% & 58.0\% & 58.0\% \\
DQC.US.0117 & 7.1\% & 7.0\% & 7.0\% \\
DQC.US.0126 & 35.1\% & 35.0\% & 35.0\% \\
\bottomrule
\end{tabular}
\caption{The evaluation set and its ablation and hard subsets. Filings and issuers are unique counts within each column; rule percentages use the corresponding instance count as the denominator.}
\label{tab:dataset}
\end{table}

\paragraph{Hard-subset selection.}
The external-agent study uses cases that require more structural processing without the audit environment. The hard subset has a median of 6 dimensional slices and 3 calculation children per instance, compared with 0 and 2 in the full evaluation set. Its seven DQC.US.0117 cases have a median of 72 dimensional slices and 37.8\,MB of XBRL per instance. This subset tests the usefulness of the environment on structurally demanding filings; it is not a random sample for estimating performance across the full evaluation set.

Table~\ref{tab:dataset-fields} gives the instance schema. The reference rule and answer are retained for scoring and are not supplied to the evaluated systems. Source-round and original identifiers preserve each instance's provenance.

\begin{table}[t]
\centering
\small
\setlength{\tabcolsep}{4pt}
\resizebox{\linewidth}{!}{
\begin{tabular}{ll}
\toprule
Field & Description \\
\midrule
\texttt{id} & Instance identifier. \\
\texttt{query} & Natural-language audit query. \\
\texttt{filing\_name} & Filing type, such as 10-K or 10-Q. \\
\texttt{dqc\_rule} & DQC rule identifier. \\
\texttt{ticker} & Company ticker used to retrieve the filing. \\
\texttt{issue\_time} & Filing release date. \\
\texttt{usgaap\_concept} & Target taxonomy concept to audit. \\
\texttt{period} & Reporting period of the target fact. \\
\texttt{gt\_answer} & Ground-truth reported value and expected value. \\
\texttt{source} & Collection round. \\
\texttt{orig\_id} & Identifier in the source round. \\
\texttt{orig\_merged\_id} & Identifier in the merged collection. \\
\bottomrule
\end{tabular}
}
\caption{Schema of the audit instances used in our experiments.}
\label{tab:dataset-fields}
\end{table}

For evaluation, we use the instance metadata to identify the target ticker and audit query, then retrieve the corresponding raw XBRL filing package from SEC EDGAR. We also parse the relevant US-GAAP taxonomy release independently. The raw filing is used to build the dynamic filing graph, while the taxonomy release is used to build the static regulatory graph. This setup avoids evaluating only on pre-extracted textual evidence and requires the system to perform verification over structured filing and taxonomy artifacts.

\section{Model Details}
\label{app:models}

Table~\ref{tab:models} gives the model identifiers used in the experiment configurations. Qwen3.6-27B and Fino1-14B are served locally with \texttt{sglang}; the other backbones use hosted endpoints, including the open-weight Qwen3.5-397B-A17B. The judge is fixed across methods and backbones.

\begin{table}[t]
\centering
\small
\setlength{\tabcolsep}{3pt}
\resizebox{\linewidth}{!}{%
\begin{tabular}{llr}
\toprule
Backbone & API name / repository & Scale \\
\midrule
Claude Sonnet 4.6 & \texttt{claude-sonnet-4-6} & undisclosed \\
GPT-4o            & \texttt{gpt-4o} & undisclosed \\
GPT-6.1-sol       & \texttt{gpt-6.1-sol} & undisclosed \\
Qwen3.5-397B-A17B & \texttt{qwen/qwen3.5-397b-a17b} & 397B \\
Qwen3.6-27B       & \texttt{Qwen/Qwen3.6-27B} & 27B \\
Fino1-14B         & \texttt{TheFinAI/Fino1-14B} & 14B \\
\midrule
Judge             & \texttt{gpt-5-mini} & undisclosed \\
\bottomrule
\end{tabular}}
\caption{Model identifiers used in the experiments. Scale denotes total parameter count where reported; undisclosed scales are not estimated.}
\label{tab:models}
\end{table}

\section{Evaluation Metric Details}
\label{app:metrics}

This appendix defines the audit metrics, judge protocol, tool-use statistics, and inter-agent disagreement measures used in the experiments.

\paragraph{LLM-as-a-judge protocol.}
For case $i$, the full artifact is $z_i=(\hat{y}_i,\hat{v}^{\mathrm{rep}}_i,\hat{v}^{\mathrm{exp}}_i,\hat{\mathcal{R}}_i,\hat{\mathcal{P}}_i,\hat{\mathcal{Z}}_i,\tau_i)$, using the fields defined in Section~\ref{sec:task}. Only the translated \texttt{extracted\_value} and \texttt{calculated\_value} fields are passed to the judge.
Following FinAuditing~\citep{10.1145/3805712.3808578}, an LLM-as-a-judge compares the reported and expected values with the reference and assigns one mutually exclusive label:
\emph{Accurate} (A), \emph{Structural} (S), \emph{Extraction} (E), or \emph{Calculation} (C).
The judge first checks structure, then the reported value, then the expected value, comparing numeric meaning rather than string formatting and requiring exact numerical equality. A case receives \emph{Accurate} only if all three checks pass. A \emph{Structural} error means the output fails the required two-field JSON structure check. An \emph{Extraction} error means the system recovers the reported fact or reported value incorrectly. A \emph{Calculation} error means the system reaches a valid structure and extracts the reported value correctly, but produces an incorrect expected value.

\paragraph{Joint audit accuracy.}
The main metric is joint audit accuracy:
\begin{equation}
  \mathrm{ACC}
  = \frac{1}{N}\sum_{i=1}^{N}\mathbf{1}[\ell_i=\mathrm{A}],
\end{equation}
where $\ell_i$ is the judge label for case $i$. The judge checks the two numerical fields, not the verdict, rule identifiers, rationale, evidence trail, or aggregation score. Across the six main \textsc{AuditFlow} runs, every A-labeled output also has the correct binary verdict. VAcc separately scores that verdict, while Joint ACC requires correct reported and expected values.

\paragraph{Rule-dependent value selection.}
Rule attribution is not independently scored. In the main protocol, the translator selects the expected value using the system's rules, with priority DQC.US.0126, DQC.US.0015, then DQC.US.0117 when a candidate is available. The respective candidates are the calculation-tree sum, the sign-corrected reported value, and the selected dimensional sum. Incorrect attribution can therefore affect Joint ACC through value selection. Appendix~\ref{app:protocol} supplies the reference rule only as a separate oracle control.

\paragraph{Per-rule accuracy.}
We also compute Joint ACC for each DQC rule family:
\begin{equation}
  \mathrm{ACC}_{r}
  = \frac{1}{|\mathcal{C}_{r}|}
    \sum_{i\in\mathcal{C}_{r}}\mathbf{1}[\ell_i=\mathrm{A}],
\end{equation}
where $\mathcal{C}_{r}$ is the subset of cases governed by rule family
$r\in\{\mathrm{0015},\mathrm{0117},\mathrm{0126}\}$. This breakdown is useful because the three rule families stress different capabilities: sign metadata lookup, dimensional aggregation, and calculation-linkbase arithmetic.

\paragraph{Verdict-only accuracy.}
To separate decision quality from numerical artifact quality, we report verdict-only accuracy:
\begin{equation}
  \mathrm{VAcc}
  = \frac{1}{N}\sum_{i=1}^{N}\mathbf{1}[\hat{y}_i=y_i].
\end{equation}
VAcc ignores numerical reconstruction errors. Since all evaluated cases are engine-recorded violations, it measures recognition of these cases and provides no estimate of specificity on compliant filings.

\paragraph{Diagnostic error rates.}
We report the three non-accurate judge-label rates:
\begin{equation}
  \mathrm{Err}_{q}
  = \frac{1}{N}\sum_{i=1}^{N}\mathbf{1}[\ell_i=q],
  \qquad q\in\{\mathrm{S},\mathrm{E},\mathrm{C}\}.
\end{equation}
These correspond to structural error rate (SER), extraction error rate (EER), and calculation error rate (CER). They form mutually exclusive diagnostic partitions, not independent failure probabilities: a case failing extraction is not also counted as a calculation error.

\paragraph{Grounded interaction metrics.}
For the agentic process, we report mean tool calls per case, required-tool gate rejections, and the shares of static and dynamic tool calls. We also measure the dynamic-tool share at each tool step. These statistics describe how each backbone implements the role-specific search policies; the tool catalogs and mandatory checks constrain the observed behavior.

\paragraph{Inter-agent disagreement.}
We report the ER conflict mass $K$ between the two junior reports. High conflict means that the last recorded junior reports assign support to opposing verdicts before final fusion; disagreement groups compare their binary verdicts. We examine whether disagreement is associated with lower final accuracy, without assuming that conflict is a calibrated difficulty score.

\begin{table*}[t]
\centering
\small
\setlength{\tabcolsep}{4pt}
\begin{tabular}{p{0.18\linewidth} p{0.20\linewidth} p{0.52\linewidth}}
\toprule
Metric family & Metric & What it measures \\
\midrule
Audit artifact &
Joint ACC &
Percentage of cases labeled \emph{Accurate} by the judge; requires a valid audit artifact with correct reported and expected values. \\
Audit artifact &
Per-rule ACC &
Joint ACC computed separately for DQC.US.0015, DQC.US.0117, and DQC.US.0126. \\
Audit artifact &
Verdict-only ACC &
Recognition of known violations in this evaluation, ignoring numerical errors; does not measure specificity. \\
Error diagnosis &
SER &
Structural error rate; failure of the required two-field JSON structure check. \\
Error diagnosis &
EER &
Extraction error rate; incorrect reported fact or reported value recovered from the filing. \\
Error diagnosis &
CER &
Calculation error rate; valid structure and correct extraction, but incorrect expected value. \\
Tool-use process &
Tool calls/case; gate rejections &
Execution statistics under the ReAct protocol and required-tool gate. \\
Environment trajectory &
Static share; dynamic share by step &
Use of static and dynamic tool catalogs under assigned policies; these are not exclusive graph-access categories. \\
Inter-agent evidence &
Conflict mass $K$ &
Degree of disagreement between junior auditors before final ER fusion. \\
\bottomrule
\end{tabular}
\caption{Full evaluation metrics used in the experiments. Audit metrics score numerical artifacts or binary verdicts; interaction metrics describe execution under assigned tool constraints.}
\label{tab:metrics}
\end{table*}

\section{Implementation Details}
\label{app:implementation}

\paragraph{Software stack.}
\textsc{AuditFlow} is implemented in Python as an executable audit pipeline rather than a prompt-only benchmark. The graph layer is built with \texttt{networkx}\footnote{\url{https://networkx.org/}}; XBRL filings and taxonomy linkbases are parsed with \texttt{lxml}\footnote{\url{https://lxml.de/}}; and tabular filing artifacts are handled with \texttt{openpyxl}\footnote{\url{https://openpyxl.readthedocs.io/}}. The agent layer uses \texttt{langgraph} and \texttt{langchain} abstractions, with provider adapters for OpenAI, Anthropic, Together, Fireworks, OpenRouter, local OpenAI-compatible servers, and HuggingFace Transformers. The implementation keeps the audit environment, tool interface, LLM backend, evidential aggregation, and evaluation code separate, so changing a backbone does not change the graph construction, deterministic checkers, ER fusion, or judge pipeline.

\paragraph{Model configuration.}
Unless otherwise stated, a single backbone is used uniformly for all three agents in a run: the Compliance Auditor $A_1$, the Forensic Auditor $A_2$, and the Senior Auditor $A_s$. For compatible standard models, we use temperature $0$. Junior output budgets are $4096$ tokens for Sonnet, Qwen3.5-397B-A17B, GPT-6.1-sol, and Qwen3.6-27B, $2048$ for GPT-4o, and $3072$ for Fino1-14B. The senior auditor receives an $8192$-token budget because it reads both junior reports. For OpenAI reasoning-family models, we use the provider's reasoning interface, with \texttt{reasoning\_effort=low} for audit agents when applicable, and do not pass temperature parameters. The offline LLM-as-a-judge is fixed across all runs as \texttt{gpt-5-mini} with \texttt{reasoning\_effort=minimal}, \texttt{verbosity=low}, and a $256$-token output budget.

\paragraph{Interaction budgets.}
Each junior invocation has a budget of $20$ ReAct steps, including tool calls, rejected terminal actions, and malformed actions. At exhaustion, the loop returns \texttt{budget\_exhausted}, constructing a report from available observations without executing missing tools. It retains observed positive rule findings and available values; absent positive findings yield an initial consistent verdict, while unresolved values may remain null. Subsequent report assembly and aggregation proceed normally. The status is recorded in the trajectory; the resulting numerical output is scored by the same judge, not automatically treated as successfully verified or assigned a structural error solely because the budget was exhausted.

The required-tool gate accepts dictionary observations without an \texttt{error} field. This is weaker than conclusive verification: \texttt{no\_data} and an \texttt{ok} calculation result with unresolved expected value can satisfy the gate. The calculation checker has no separate \texttt{not\_evaluable} status (Appendix~\ref{app:graph-construction}).

The initial junior investigation can be followed by at most $3$ refinement rounds, each with one tool-free senior feedback call. Single-junior ablations omit this loop. Main evaluations use up to $8$ concurrent cases.

\paragraph{Local and remote inference.}
Qwen3.6-27B and Fino1-14B are served through local OpenAI-compatible \texttt{sglang} endpoints, with context windows of 131{,}072 and 32{,}768 tokens, respectively. Qwen's thinking mode is disabled in the chat template. Fino1 runs without a reasoning parser: enabling that parser moves its generated content out of the response body expected by the agent loop. The reported Fino1 results use the corrected configuration. Three instances exceed its context window and count as structural failures in the 1{,}000-instance evaluation; its trajectory analyses use the 997 completed cases. The remaining backbones use hosted APIs.

\paragraph{Reproducibility and caching.}
The pipeline writes one JSONL audit artifact per run and a companion trajectory sidecar containing per-agent tool calls, observations, parse failures, gate rejections, and finalization status. LLM calls are cached on disk with keys that include the role, model name, system prompt, canonical message history, and sorted tool names. The judge is cached separately. Replaying an unchanged run against a warm cache reproduces the same audit artifacts and judge labels without additional LLM calls.

\section{Full Experimental Results}
\label{app:full-results}

Tables~\ref{tab:full} and~\ref{tab:full-cont} report all main method--backbone combinations on the same 1{,}000 instances. Joint ACC, SER, EER, and CER partition the evaluation set. Methods are shown as group headings, with one row per backbone.

\begin{table}[t]
\centering
\footnotesize
\setlength{\tabcolsep}{3pt}
\begin{tabular}{@{}lrrrr@{}}
\toprule
Backbone & Joint ACC $\uparrow$ & SER $\downarrow$ & EER $\downarrow$ & CER $\downarrow$ \\
\midrule
\multicolumn{5}{@{}l}{\textit{Checker-only}} \\
No LLM & 86.70 & 4.50 & 0.70 & 8.10 \\
\midrule
\multicolumn{5}{@{}l}{\textit{FinAuditing}} \\
Sonnet 4.6 & 7.80 & 0.10 & 21.90 & 70.20 \\
GPT-4o & 6.20 & 0.10 & 19.80 & 73.90 \\
Qwen3.5-397B & 6.40 & 1.60 & 21.10 & 70.90 \\
GPT-6.1-sol & 15.20 & 0.10 & 16.10 & 68.60 \\
Qwen3.6-27B & 4.70 & 14.40 & 29.50 & 51.40 \\
Fino1-14B & 14.60 & 69.50 & 12.50 & 3.40 \\
\midrule
\multicolumn{5}{@{}l}{\textit{Direct LLM}} \\
Sonnet 4.6 & 1.70 & 87.40 & 0.20 & 10.70 \\
GPT-4o & 6.20 & 59.10 & 3.20 & 31.50 \\
Qwen3.5-397B & 9.20 & 53.20 & 1.60 & 36.00 \\
GPT-6.1-sol & 7.80 & 59.10 & 3.10 & 30.00 \\
Qwen3.6-27B & 1.70 & 63.60 & 1.70 & 33.00 \\
Fino1-14B & 7.80 & 56.10 & 9.20 & 26.90 \\
\midrule
\multicolumn{5}{@{}l}{\textit{Vanilla RAG}} \\
Sonnet 4.6 & 3.20 & 96.10 & 0.50 & 0.20 \\
GPT-4o & 1.70 & 88.80 & 7.90 & 1.60 \\
Qwen3.5-397B & 4.70 & 94.50 & 0.60 & 0.20 \\
GPT-6.1-sol & 7.80 & 91.50 & 0.40 & 0.30 \\
Qwen3.6-27B & 7.70 & 91.30 & 0.80 & 0.20 \\
Fino1-14B & 1.70 & 93.10 & 1.90 & 3.30 \\
\midrule
\multicolumn{5}{@{}l}{\textit{GraphRAG}} \\
Sonnet 4.6 & 46.80 & 11.30 & 0.60 & 41.30 \\
GPT-4o & 39.20 & 17.50 & 6.30 & 37.00 \\
Qwen3.5-397B & 27.40 & 38.20 & 1.90 & 32.50 \\
GPT-6.1-sol & 49.60 & 4.20 & 1.80 & 44.40 \\
Qwen3.6-27B & 22.80 & 44.10 & 4.90 & 28.20 \\
Fino1-14B & 18.30 & 62.00 & 15.20 & 4.50 \\
\bottomrule
\end{tabular}
\caption{Full main results (\%): deterministic, direct, and retrieval methods. Every row uses 1{,}000 instances. Remaining methods appear in Table~\ref{tab:full-cont}.}
\label{tab:full}
\end{table}

\begin{table}[t]
\centering
\footnotesize
\setlength{\tabcolsep}{3pt}
\begin{tabular}{@{}lrrrr@{}}
\toprule
Backbone & Joint ACC $\uparrow$ & SER $\downarrow$ & EER $\downarrow$ & CER $\downarrow$ \\
\midrule
\multicolumn{5}{@{}l}{\textit{TreeRAG}} \\
Sonnet 4.6 & 42.20 & 33.90 & 1.80 & 22.10 \\
GPT-4o & 36.20 & 20.50 & 1.80 & 41.50 \\
Qwen3.5-397B & 33.30 & 38.20 & 7.80 & 20.70 \\
GPT-6.1-sol & 43.70 & 32.30 & 3.20 & 20.80 \\
Qwen3.6-27B & 24.20 & 36.80 & 3.30 & 35.70 \\
Fino1-14B & 21.30 & 44.20 & 7.80 & 26.70 \\
\midrule
\multicolumn{5}{@{}l}{\textit{Herculean}} \\
Sonnet 4.6 & 65.90 & 0.10 & 0.40 & 33.60 \\
GPT-4o & 66.10 & 1.70 & 9.20 & 23.00 \\
Qwen3.5-397B & 43.70 & 0.20 & 12.30 & 43.80 \\
GPT-6.1-sol & 66.20 & 0.10 & 6.20 & 27.50 \\
Qwen3.6-27B & 37.70 & 3.20 & 12.20 & 46.90 \\
Fino1-14B & 28.70 & 42.80 & 22.60 & 5.90 \\
\midrule
\multicolumn{5}{@{}l}{\textit{Single Agent}} \\
Sonnet 4.6 & 68.60 & 19.10 & 2.90 & 9.40 \\
GPT-4o & 68.80 & 19.60 & 2.80 & 8.80 \\
Qwen3.5-397B & 68.50 & 19.60 & 2.60 & 9.30 \\
GPT-6.1-sol & 69.10 & 21.00 & 1.00 & 8.90 \\
Qwen3.6-27B & 66.30 & 19.50 & 2.40 & 11.80 \\
Fino1-14B & 58.10 & 23.90 & 0.90 & 17.10 \\
\midrule
\multicolumn{5}{@{}l}{\textsc{AuditFlow}} \\
Sonnet 4.6 & 87.70 & 3.60 & 0.80 & 7.90 \\
GPT-4o & 87.00 & 4.20 & 0.70 & 8.10 \\
Qwen3.5-397B & 86.70 & 4.40 & 0.70 & 8.20 \\
GPT-6.1-sol & 84.00 & 6.30 & 1.40 & 8.30 \\
Qwen3.6-27B & 78.00 & 10.30 & 1.40 & 10.30 \\
Fino1-14B & 83.90 & 5.60 & 0.90 & 9.60 \\
\bottomrule
\end{tabular}
\caption{Full main results (\%), continued: TreeRAG, agentic baselines, and \textsc{AuditFlow}. Every row uses 1{,}000 instances, including Fino1's three context-window failures as structural errors.}
\label{tab:full-cont}
\end{table}

\setlength{\floatsep}{6pt plus 2pt minus 2pt}
\setlength{\textfloatsep}{10pt plus 2pt minus 2pt}
\setlength{\intextsep}{8pt plus 2pt minus 2pt}

\subsection{Verdict and Artifact Accuracy}
\label{app:backbone}

\begin{table}[!htbp]
\centering
\small
\setlength{\tabcolsep}{3pt}
\resizebox{\linewidth}{!}{%
\begin{tabular}{lrrr}
\toprule
Backbone & Joint ACC $\uparrow$ & VAcc $\uparrow$ & Gap \\
\midrule
Claude Sonnet 4.6    & 87.70 & 94.60 & 6.90 \\
GPT-4o               & 87.00 & 93.80 & 6.80 \\
Qwen3.5-397B-A17B    & 86.70 & 93.50 & 6.80 \\
GPT-6.1-sol          & 84.00 & 94.80 & 10.80 \\
Qwen3.6-27B          & 78.00 & 90.60 & 12.60 \\
Fino1-14B            & 83.90 & 91.00 & 7.10 \\
\bottomrule
\end{tabular}}
\caption{Verdict-only accuracy and Joint ACC on 1{,}000 known violations. Gap is VAcc minus Joint ACC, in percentage points. VAcc tests neither numerical reconstruction nor specificity on compliant filings.}
\label{tab:vacc}
\end{table}

Sonnet, GPT-4o, and Qwen3.5-397B-A17B match Checker-only's $66.20\%$ accuracy on DQC.US.0117. The other backbones score lower, so dimensional performance is not invariant to the agent. The common result among the strongest three backbones suggests that improving the dimensional checker is an important remaining direction, while the lower results show that an agent can also fail to preserve the checker's accuracy.

\subsection{Ablation Diagnostics}
\label{app:ablation}

The full configurations in Table~\ref{tab:ablation} score $83.33\%$ on GPT-6.1-sol and $88.67\%$ on Sonnet on the 300-instance sample. These subset scores should not be compared directly with the 1{,}000-instance main scores. Removing the gate lowers Joint ACC on GPT-6.1-sol, while leaving Sonnet unchanged.

The single-$A_1$ variant removes the Forensic Auditor; the single-$A_2$ variant removes the Compliance Auditor. On GPT-6.1-sol, their Joint ACC losses are $4.67$ and $12.67$ points; on Sonnet, they are $0.33$ and $19.00$ points. Both juniors retain access to deterministic checkers, so the difference cannot be explained by checker access alone. Removing ER fusion costs $2.67$ points on GPT-6.1-sol and leaves Sonnet unchanged, consistent with the backbone-dependent effects in Section~\ref{sec:ablation}.

\section{Checker Coverage and Reporting Protocol}
\label{app:coverage}

\subsection{Coverage and Override Permission}

The limited sign checker uses a curated concept list and a debit-balance fallback; credit-balance concepts are exempt because some legitimately admit negatives. Of the 274 official-engine violations missed on the 1{,}000-instance set, 258 have a credit balance. Expanded coverage includes those concepts at a potential precision cost, without reproducing the official engine. Authoritative reporting follows checker outcomes; override-enabled reporting permits declared, evidence-supported additions while forbidding removal of positive findings. These vary coverage and agent permission separately.

Table~\ref{tab:coverage} in Section~\ref{sec:coverage} reports all conditions on the same 300 instances from the ablation subset.

Expanding coverage under authoritative reporting adds $17.00$ points for Sonnet and $24.00$ for GPT-4o. Limited-coverage overrides recover $14.67$ and $11.00$ points, approximately $86\%$ and $46\%$ of those gaps. Qwen3.6-27B peaks at $80.00\%$ with expanded coverage and authoritative reporting, illustrating the backbone dependence of overrides.

\subsection{Effect of the Reference Rule Label}
\label{app:protocol}

The standard translator uses system-identified rules. This oracle control supplies the reference rule and selects its numerical candidate even without a positive checker finding. Table~\ref{tab:protocol} holds coverage limited and varies output selection, separately from checker coverage and override permission. These scores do not replace the main expanded-coverage results.

\begin{table}[!htbp]
\centering
\small
\setlength{\tabcolsep}{3pt}
\resizebox{\linewidth}{!}{%
\begin{tabular}{lrrr}
\toprule
System & Own rules & GT label & $\Delta$ \\
\midrule
Claude Sonnet 4.6  & 64.70 & 79.90 & $+15.20$ \\
Qwen3.6-27B        & 62.60 & 74.10 & $+11.50$ \\
GPT-4o             & 60.80 & 78.30 & $+17.50$ \\
Qwen3.5-397B-A17B  & 59.10 & 78.30 & $+19.20$ \\
GPT-6.1-sol        & 58.00 & 79.90 & $+21.90$ \\
Checker-only       & 58.80 & 77.90 & $+19.10$ \\
\bottomrule
\end{tabular}}
\caption{Joint ACC (\%) under limited checker coverage on the same 1{,}000 instances. Own rules uses system-selected attribution; GT label supplies the reference rule to the output translator. The latter is an oracle control and is not used for headline results. Fino1-14B is omitted because its control runs predate the corrected serving configuration.}
\label{tab:protocol}
\end{table}

The reference label adds $11.50$--$21.90$ points. GPT-6.1-sol gains most ($58.00\%$ to $79.90\%$); its own-rule score is below Checker-only's $58.80\%$. Checker-only itself gains $19.10$ points without an LLM. This is an effect of rule-dependent value selection, not improved model reasoning, motivating exclusion of reference labels from the main protocol.

\section{External Coding-Agent Evaluation}
\label{app:external}

We evaluate Claude Code~\citep{anthropic2026claudecode}, Codex~\citep{openai2026codexcli}, OpenClaw~\citep{openclaw2026}, and DeepSeek Harness~\citep{deepseek2026harness}, paired with the models listed in Table~\ref{tab:external}. Within each row, \textsc{bare} and \textsc{env} use the same harness and model.

Both conditions receive the filing's XBRL files, question, and three rule descriptions, and can write parsing code; \textsc{env} additionally provides command-line access to the environment. Runs use \texttt{bubblewrap}, with outbound access limited to the model API through an allow-listed proxy. Web tools are disabled and reference files are not mounted.

The 100 difficulty-selected cases comprise 58 sign, 7 dimensional, and 35 calculation-tree cases. Scoring uses exact matches for \texttt{extracted\_value} and \texttt{calculated\_value}. Win/Lose count corrections/regressions; Tie includes joint successes and failures. Two-sided exact McNemar tests are unadjusted. Runs used a separate machine from the main evaluation.

Model and harness vary together across rows, preventing attribution of gain differences to the backbone alone. Neither condition solves the seven dimensional cases.

\section{Agent-Behavior Details}
\label{app:agent-behavior}

\subsection{Tool Use and Search Order}
\label{app:tool-use}

\begin{table}[!htbp]
\centering
\small
\setlength{\tabcolsep}{3pt}
\resizebox{\linewidth}{!}{%
\begin{tabular}{lrrrr}
\toprule
\multirow{2}{*}{Backbone} & \multicolumn{2}{c}{Calls per case} & \multicolumn{2}{c}{$A_2$ tool mix} \\
\cmidrule(lr){2-3}\cmidrule(lr){4-5}
 & $A_1$ & $A_2$ & Static & Dynamic \\
\midrule
Claude Sonnet 4.6    & 5.03 & 5.92 & 61.3\% & 38.7\% \\
GPT-4o               & 4.99 & 5.44 & 55.2\% & 44.8\% \\
Qwen3.5-397B-A17B    & 4.01 & 5.53 & 58.8\% & 41.2\% \\
GPT-6.1-sol          & 7.99 & 5.69 & 60.8\% & 39.2\% \\
Qwen3.6-27B          & 4.61 & 4.58 & 69.9\% & 30.1\% \\
Fino1-14B            & 3.85 & 8.34 & 65.7\% & 34.3\% \\
\bottomrule
\end{tabular}}
\caption{Mean calls and $A_2$'s tool-catalog mix over 1{,}000 completed trajectories, except Fino1 (997). Its three observation overflows omit both junior traces; payloads are at or above the 99.8th percentile. Other backbones complete these cases with near-average call counts, which does not establish missing Fino1 counts. Means are conditional on completion.}
\label{tab:tooluse}
\end{table}

GPT-6.1-sol makes the most $A_1$ calls ($7.99$ per case), and Fino1 the most $A_2$ calls ($8.34$). These counts describe constrained search behavior, not the benefit of additional calls.

\begin{table}[!htbp]
\centering
\small
\setlength{\tabcolsep}{3pt}
\resizebox{\linewidth}{!}{%
\begin{tabular}{lrrrrrr}
\toprule
Backbone & $k{=}0$ & $k{=}1$ & $k{=}2$ & $k{=}3$ & $k{=}4$ & $k{=}5$ \\
\midrule
Claude Sonnet 4.6    & 100 & 75 & 24 & 14 & 5 & 6 \\
GPT-4o               & 100 & 100 & 22 & 22 & 1 & 0 \\
Qwen3.5-397B-A17B    & 83 & 99 & 32 & 11 & 2 & 1 \\
GPT-6.1-sol          & 100 & 96 & 23 & 4 & 0 & 0 \\
Qwen3.6-27B          & 100 & 25 & 8 & 4 & 1 & 0 \\
Fino1-14B            & 29 & 20 & 20 & 42 & 67 & 53 \\
\bottomrule
\end{tabular}}
\caption{Share (\%) of $A_2$ calls using dynamic filing tools at zero-indexed step $k$. Steps are reported while at least 10 instances reach them. Later steps are conditional on continuing trajectories, so the columns need not contain the same set of instances.}
\label{tab:switchover}
\end{table}

Five backbones begin with dynamic filing tools and later use taxonomy tools and checks; Fino1's dynamic share rises later. Catalog access and mandatory calls constrain these patterns.

\paragraph{Role specialization control.}
The main experiments restrict $A_1$ to the static catalog, so its static-call share alone cannot demonstrate a preference. We therefore add the nine dynamic filing tools available to $A_2$, keeping $A_1$'s persona, required checks, protocol configuration, and GPT-6.1-sol backbone fixed. This is a separate control; the main catalogs remain unchanged. The static-only reference is the full-system configuration in Table~\ref{tab:ablation}, evaluated on the same 300 instance identifiers.

Table~\ref{tab:role-specialization-control} in Section~\ref{sec:agent-behavior} shows little change in call volume or system accuracy. With both catalogs, $A_1$ makes four dynamic calls among 2{,}384 calls, each in a different instance. The count of \texttt{walk\_calc\_tree} calls changes from 592 to 586; \texttt{lookup\_concept\_spec}, \texttt{check\_sign}, \texttt{check\_calc\_tree}, and \texttt{check\_\allowbreak dim\_\allowbreak consistency} each remain at 300 calls. Joint ACC changes from $83.33\%$ to $84.00\%$, a net difference of two correct cases, not evidence of an accuracy benefit. The static-tool preference persists without the catalog restriction, but tool descriptions, mandatory checks, and backbone preferences remain possible influences. This control neither isolates the persona's causal effect nor establishes that multiple auditors are necessary.

\subsection{Inter-Agent Disagreement}
\label{app:inter-agent-disagreement}

\begin{table}[!htbp]
\centering
\small
\setlength{\tabcolsep}{3pt}
\resizebox{\linewidth}{!}{%
\begin{tabular}{llrrr}
\toprule
Backbone & Junior reports & Cases & Final ACC & Mean $K$ \\
\midrule
Claude Sonnet 4.6    & Agree     & 992 & 88.0\% & 0.000 \\
Claude Sonnet 4.6    & Disagree  & 8 & 50.0\% & 0.647 \\
\addlinespace
GPT-4o               & Agree     & 994 & 87.2\% & 0.000 \\
GPT-4o               & Disagree  & 6 & 50.0\% & 0.594 \\
\addlinespace
Qwen3.5-397B-A17B    & Agree     & 999 & 86.7\% & 0.000 \\
Qwen3.5-397B-A17B    & Disagree  & 1 & 100.0\% & 0.808 \\
\addlinespace
GPT-6.1-sol          & Agree     & 991 & 84.5\% & 0.000 \\
GPT-6.1-sol          & Disagree  & 9 & 33.3\% & 0.665 \\
\addlinespace
Qwen3.6-27B          & Agree     & 582 & 79.2\% & 0.000 \\
Qwen3.6-27B          & Disagree  & 418 & 76.3\% & 0.770 \\
\addlinespace
Fino1-14B            & Agree     & 989 & 84.2\% & 0.000 \\
Fino1-14B            & Disagree  & 8 & 75.0\% & 0.820 \\
\bottomrule
\end{tabular}}
\caption{Final accuracy grouped by verdict agreement in the last recorded junior reports. There are 1{,}000 cases per backbone except Fino1-14B, whose 997 completed cases appear here; its three context failures remain in all main accuracy denominators. Mean $K$ uses the mass mapping in Appendix~\ref{app:fusion}, consistent with the saved fused distributions.}
\label{tab:disagreement}
\end{table}

Agreement groups have higher accuracy for Sonnet, GPT-4o, and GPT-6.1-sol, but each has only 6--9 disagreement cases; Qwen3.5 has just one. Qwen3.6 disagrees more often with a smaller accuracy gap. Zero $K$ on agreement follows from the mapping, not correctness. Disagreement-side $K$ depends on reported rules and value availability, not a calibrated error probability.

\section{Bootstrap Analysis for Backbone Comparisons}
\label{app:bootstrap}

We draw $B=10^4$ instance-level resamples with replacement for 95\% percentile intervals. All six backbones share sampled indices over 1{,}000 sorted case identifiers, with Fino1's three failures scored zero. Sampling uses Python's \texttt{random.Random(0).choices}.

\begin{table}[!htbp]
\centering
\small
\setlength{\tabcolsep}{3pt}
\resizebox{\linewidth}{!}{%
\begin{tabular}{lrrr}
\toprule
Backbone & Joint ACC & Correct & 95\% CI \\
\midrule
Claude Sonnet 4.6 & 87.70 & 877/1000 & [85.6, 89.7] \\
GPT-4o & 87.00 & 870/1000 & [84.9, 89.1] \\
Qwen3.5-397B-A17B & 86.70 & 867/1000 & [84.5, 88.8] \\
GPT-6.1-sol & 84.00 & 840/1000 & [81.7, 86.2] \\
Fino1-14B & 83.90 & 839/1000 & [81.6, 86.1] \\
Qwen3.6-27B & 78.00 & 780/1000 & [75.4, 80.5] \\
\bottomrule
\end{tabular}}
\caption{Joint ACC (\%) with 95\% percentile bootstrap intervals on the full 1{,}000-instance evaluation set.}
\label{tab:boot-acc}
\end{table}

\begin{table}[!htbp]
\centering
\small
\setlength{\tabcolsep}{3pt}
\resizebox{\linewidth}{!}{%
\begin{tabular}{lrrr}
\toprule
Compared with Sonnet & $\Delta$ACC & 95\% CI & $p$ \\
\midrule
GPT-4o & +0.7 & [+0.2, +1.3] & 0.016 \\
Qwen3.5-397B-A17B & +1.0 & [+0.4, +1.6] & 0.003 \\
GPT-6.1-sol & +3.7 & [+2.4, +5.0] & $<0.001$ \\
Fino1-14B & +3.8 & [+2.5, +5.2] & $<0.001$ \\
Qwen3.6-27B & +9.7 & [+7.9, +11.6] & $<0.001$ \\
\bottomrule
\end{tabular}}
\caption{Paired comparisons on the same 1{,}000 cases. $\Delta$ACC is Sonnet minus the listed backbone, in percentage points. All tests use $10^4$ shared resamples; $p$-values are unadjusted.}
\label{tab:boot-pairs}
\end{table}

Paired tests resample correctness differences and use two-sided recentred bootstrap $p$-values~\citep{berg2012empirical}. Sonnet's gains over GPT-4o and Qwen3.5 are small despite intervals excluding zero; other gaps are larger. These unadjusted, instance-level comparisons condition on this sample. They capture neither repeated-run variation nor dependence within filings or issuers, and are not cluster-robust population estimates.

\section{Human Evaluation Annotation Guidelines}
\label{app:human-annotation}

This appendix documents the complete annotation instructions provided to the two finance experts in the human evaluation described in Section~\ref{sec:metrics}.

The two co-author annotators independently assessed 100 \textsc{AuditFlow} outputs generated using Claude Sonnet 4.6. Each prediction was compared against its reference answer using four mutually exclusive categories: Accurate (A), Structural Error (S), Extraction Error (E), and Calculation Error (C).

\paragraph{Annotation task.}
For each evaluation instance, the experts were provided with a reference answer (\texttt{true\_answer}) and a model prediction (\texttt{pred\_answer}). Each answer contained two audit-critical numerical fields: \texttt{extracted\_value} and \texttt{calculated\_value}. The experts assigned one label following a sequential procedure: structure validation, extracted-value comparison, and calculated-value comparison.

\paragraph{Instructions provided to annotators.}
The following reproduces the instructions provided to both experts, including the decision rules and illustrative examples.

\begin{quote}
\small
Your task is to judge whether the model's output \texttt{pred\_answer} is correct compared to the given \texttt{true\_answer}.

Follow the rules strictly:

\medskip
\noindent
\textbf{Step 1 (Structure Check):}

Verify whether \texttt{pred\_answer} has the same structure as \texttt{true\_answer}.

The required structure is a JSON object with exactly two keys:

\begin{verbatim}
{
  "extracted_value": <value>,
  "calculated_value": <value>
}
\end{verbatim}

Minor formatting differences (e.g., line breaks, indentation, whitespace) are acceptable.

If the structure is invalid, output the label: S.

If valid, continue to Step 2.

\medskip
\noindent
\textbf{Step 2 (Extracted Value Check):}

Compare \texttt{true\_answer["extracted\_value"]} and \texttt{pred\_answer["extracted\_value"]} by their mathematical meaning, not their string form.

For example, "-1,284" and "-1284" are considered equal.

If they are not equal in numeric meaning, output the label: E.

If equal, continue to Step 3.

\medskip
\noindent
\textbf{Step 3 (Calculated Value Check):}

Compare \texttt{true\_answer["calculated\_value"]} and \texttt{pred\_answer["calculated\_value"]} strictly in numeric meaning.

They must be exactly equal (zero tolerance).

If they are not equal, output the label: C.

If equal, then everything is correct.

\medskip
\noindent
\textbf{Final Decision:}

If all three checks pass, output the label: A.

Output only one label: S, E, C, or A.

Do not explain your reasoning.

\medskip
\noindent
\begin{minipage}{\linewidth}
\textbf{Example 1:}

\begin{verbatim}
true_answer =
{
  "extracted_value": "-1,286",
  "calculated_value": "1,286"
}

pred_answer =
{
  "extracted_value": "-1286",
  "calculated_value": "0"
}

Output: C
\end{verbatim}
\end{minipage}

\medskip
\noindent
\begin{minipage}{\linewidth}
\textbf{Example 2:}

\begin{verbatim}
true_answer =
{
  "extracted_value": "5,000",
  "calculated_value": "5,000"
}

pred_answer =
{
  "extracted_value": "5000",
  "calculated_value": "5000"
}

Output: A
\end{verbatim}
\end{minipage}

\medskip
\noindent
\begin{minipage}{\linewidth}
\textbf{Example 3:}

\begin{verbatim}
true_answer =
{
  "extracted_value": "123",
  "calculated_value": "456"
}

pred_answer =
{
  "extracted_value": "124",
  "calculated_value": "456"
}

Output: E
\end{verbatim}
\end{minipage}

\medskip
\noindent
\begin{minipage}{\linewidth}
\textbf{Example 4:}

\begin{verbatim}
true_answer =
{
  "extracted_value": "100",
  "calculated_value": "200"
}

pred_answer =
{
  "wrong_key": "100",
  "calculate_value": "200"
}

Output: S
\end{verbatim}
\end{minipage}
\end{quote}

\paragraph{Label interpretation.}
The four categories are mutually exclusive and assigned sequentially. A Structural Error (S) is assigned when the prediction does not satisfy the required JSON structure. If the structure is valid, an incorrect extracted value results in an Extraction Error (E). A Calculation Error (C) is assigned only when the extracted value is correct but the calculated value differs from the reference. An output is labeled Accurate (A) only when the structure and both numerical values are correct. The procedure is consistent with the evaluation categories and metric definitions in Appendix~\ref{app:metrics}.

\end{document}